# Recent Advances in Convolutional Neural Networks

Jiuxiang Gu[a,*], Zhenhua Wang[b,*], Jason Kuen[b], Lianyang Ma[b], Amir Shahroudy[b], Bing Shuai[b], Ting Liu[b], Xingxing Wang[b], Li Wang[b], Gang Wang[b], Jianfei Cai[c], Tsuhan Chen[c]

[a]*ROSE Lab, Interdisciplinary Graduate School, Nanyang Technological University, Singapore*
[b]*School of Electrical and Electronic Engineering, Nanyang Technological University, Singapore*
[c]*School of Computer Science and Engineering, Nanyang Technological University, Singapore*

## Abstract

In the last few years, deep learning has led to very good performance on a variety of problems, such as visual recognition, speech recognition and natural language processing. Among different types of deep neural networks, convolutional neural networks have been most extensively studied. Leveraging on the rapid growth in the amount of the annotated data and the great improvements in the strengths of graphics processor units, the research on convolutional neural networks has been emerged swiftly and achieved state-of-the-art results on various tasks. In this paper, we provide a broad survey of the recent advances in convolutional neural networks. We detailize the improvements of CNN on different aspects, including layer design, activation function, loss function, regularization, optimization and fast computation. Besides, we also introduce various applications of convolutional neural networks in computer vision, speech and natural language processing.

## 1. Introduction

Convolutional Neural Network (CNN) is a well-known deep learning architecture inspired by the natural visual perception mechanism of the living creatures. In 1959, Hubel & Wiesel [1] found that cells in animal visual cortex are responsible for detecting light in receptive fields. Inspired by this discovery, Kunihiko Fukushima proposed the neocognitron in 1980 [2], which could be regarded as the predecessor of CNN. In 1990, LeCun *et al.* [3] published the seminal paper establishing the modern framework of CNN, and later improved it in [4]. They developed a multi-layer artificial neural network called LeNet-5 which could classify handwritten digits. Like other neural networks, LeNet-5 has multiple layers and can be trained with the backpropagation algorithm [5]. It can obtain effective representations of the original image, which makes it possible to recognize visual patterns directly from raw pixels with little-to-none preprocessing. A parallel study of Zhang *et al.* [6] used a shift-invariant artificial neural network (SIANN) to recognize characters from an image. However, due to the lack of large training data and computing power at that time, their networks can not perform well on more complex problems, *e.g.*, large-scale image and video classification.

Since 2006, many methods have been developed to overcome the difficulties encountered in training deep CNNs [7–10]. Most notably, Krizhevsky *et al.*proposed a classic CNN architecture and showed significant improvements upon previous methods on the image classification task. The overall architecture of their method, *i.e.*, AlexNet [8], is similar to LeNet-5 but with a deeper structure. With the success of AlexNet, many works have been proposed to improve its performance. Among them, four representative works are

---

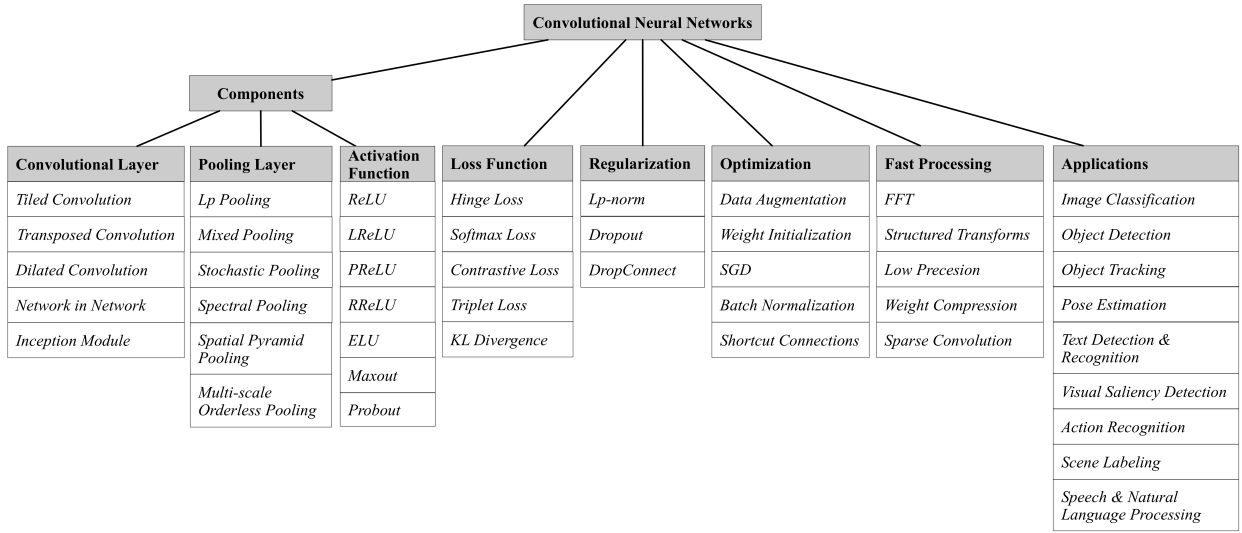

Figure 1: Hierarchically-structured taxonomy of this survey

ZFNet [11], VGGNet [9], GoogleNet [10] and ResNet [12]. From the evolution of the architectures, a typical trend is that the networks are getting deeper, e.g., ResNet, which won the champion of ILSVRC 2015, is about 20 times deeper than AlexNet and 8 times deeper than VGGNet. By increasing depth, the network can better approximate the target function with increased nonlinearity and get better feature representations. However, it also increases the complexity of the network, which makes the network be more difficult to optimize and easier to get overfitting. Along this way, various methods have been proposed to deal with these problems in various aspects. In this paper, we try to give a comprehensive review of recent advances and give some thorough discussions.

In the following sections, we identify broad categories of works related to CNN. Figure 1 shows the hierarchically-structured taxonomy of this paper. We first give an overview of the basic components of CNN in Section 2. Then, we introduce some recent improvements on different aspects of CNN including convolutional layer, pooling layer, activation function, loss function, regularization and optimization in Section 3 and introduce the fast computing techniques in Section 4. Next, we discuss some typical applications of CNN including image classification, object detection, object tracking, pose estimation, text detection and recognition, visual saliency detection, action recognition, scene labeling, speech and natural language processing in Section 5. Finally, we conclude this paper in Section 6.

## 2. Basic CNN Components

There are numerous variants of CNN architectures in the literature. However, their basic components are very similar. Taking the famous LeNet-5 as an example, it consists of three types of layers, namely convolutional, pooling, and fully-connected layers. The convolutional layer aims to learn feature representations of the inputs. As shown in Figure 2(a), convolution layer is composed of several convolution kernels which are used to compute different feature maps. Specifically, each neuron of a feature map is connected to a region of neighbouring neurons in the previous layer. Such a neighbourhood is referred to as the neuron's receptive field in the previous layer. The new feature map can be obtained by first convolving the input with a learned kernel and then applying an element-wise nonlinear activation function on the convolved results. Note that, to generate each feature map, the kernel is shared by all spatial locations of the input. The complete feature maps are obtained by using several different kernels. Mathematically, the feature value at location $(i, j)$ in the $k$-th feature map of $l$-th layer, $z^l_{i,j,k}$, is calculated by:

$$z^l_{i,j,k} = {\mathbf{w}^l_k}^T \mathbf{x}^l_{i,j} + b^l_k \tag{1}$$

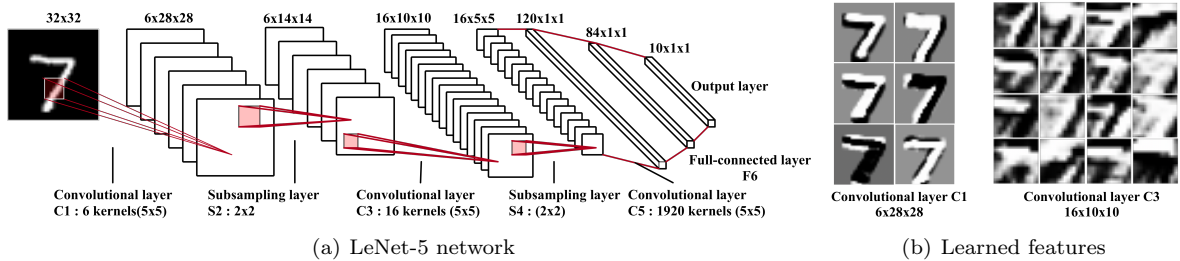

(a) LeNet-5 network  (b) Learned features

Figure 2: (a) The architecture of the LeNet-5 network, which works well on digit classification task. (b) Visualization of features in the LeNet-5 network. Each layer's feature maps are displayed in a different block.

where $\mathbf{w}_k^l$ and $b_k^l$ are the weight vector and bias term of the $k$-th filter of the $l$-th layer respectively, and $\mathbf{x}_{i,j}^l$ is the input patch centered at location $(i, j)$ of the $l$-th layer. Note that the kernel $\mathbf{w}_k^l$ that generates the feature map $\mathbf{z}_{:,:,k}^l$ is shared. Such a weight sharing mechanism has several advantages such as it can reduce the model complexity and make the network easier to train. The activation function introduces nonlinearities to CNN, which are desirable for multi-layer networks to detect nonlinear features. Let $a(\cdot)$ denote the nonlinear activation function. The activation value $a_{i,j,k}^l$ of convolutional feature $z_{i,j,k}^l$ can be computed as:

$$a_{i,j,k}^l = a(z_{i,j,k}^l) \tag{2}$$

Typical activation functions are sigmoid, tanh [13] and ReLU [14]. The pooling layer aims to achieve shift-invariance by reducing the resolution of the feature maps. It is usually placed between two convolutional layers. Each feature map of a pooling layer is connected to its corresponding feature map of the preceding convolutional layer. Denoting the pooling function as pool($\cdot$), for each feature map $\mathbf{a}_{:,:,k}^l$ we have:

$$y_{i,j,k}^l = \text{pool}(a_{m,n,k}^l), \forall (m, n) \in \mathcal{R}_{ij} \tag{3}$$

where $\mathcal{R}_{ij}$ is a local neighbourhood around location $(i, j)$. The typical pooling operations are average pooling [15] and max pooling [16]. Figure 2(b) shows the feature maps of digit 7 learned by the first two convolutional layers. The kernels in the 1st convolutional layer are designed to detect low-level features such as edges and curves, while the kernels in higher layers are learned to encode more abstract features. By stacking several convolutional and pooling layers, we could gradually extract higher-level feature representations.

After several convolutional and pooling layers, there may be one or more fully-connected layers which aim to perform high-level reasoning [9, 11, 17]. They take all neurons in the previous layer and connect them to every single neuron of current layer to generate global semantic information. Note that fully-connected layer not always necessary as it can be replaced by a $1 \times 1$ convolution layer [18].

The last layer of CNNs is an output layer. For classification tasks, the softmax operator is commonly used [8]. Another commonly used method is SVM, which can be combined with CNN features to solve different classification tasks [19, 20]. Let $\boldsymbol{\theta}$ denote all the parameters of a CNN (*e.g.*, the weight vectors and bias terms). The optimum parameters for a specific task can be obtained by minimizing an appropriate loss function defined on that task. Suppose we have $N$ desired input-output relations $\{(\boldsymbol{x}^{(n)}, \boldsymbol{y}^{(n)}); n \in [1, \cdots, N]\}$, where $\boldsymbol{x}^{(n)}$ is the $n$-th input data, $\boldsymbol{y}^{(n)}$ is its corresponding target label and $\boldsymbol{o}^{(n)}$ is the output of CNN. The loss of CNN can be calculated as follows:

$$\mathcal{L} = \frac{1}{N} \sum_{n=1}^{N} \ell(\boldsymbol{\theta}; \boldsymbol{y}^{(n)}, \boldsymbol{o}^{(n)}) \tag{4}$$

Training CNN is a problem of global optimization. By minimizing the loss function, we can find the best fitting set of parameters. Stochastic gradient descent is a common solution for optimizing CNN network [21, 22].

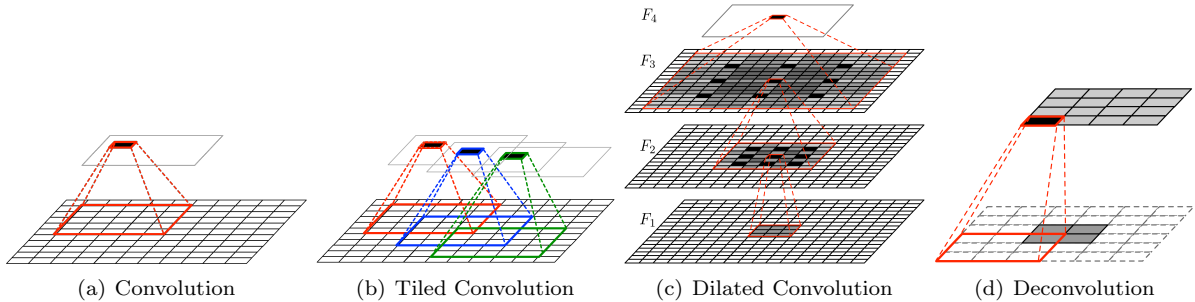

Figure 3: Illustration of (a) Convolution, (b) Tiled Convolution, (c) Dilated Convolution, and (d) Deconvolution

## 3. Improvements on CNNs

There have been various improvements on CNNs since the success of AlexNet in 2012. In this section, we describe the major improvements on CNNs from six aspects: convolutional layer, pooling layer, activation function, loss function, regularization, and optimization.

### 3.1. Convolutional Layer

Convolution filter in basic CNNs is a generalized linear model (GLM) for the underlying local image patch. It works well for abstraction when instances of latent concepts are linearly separable. Here we introduce some works which aim to enhance its representation ability.

#### 3.1.1. Tiled Convolution

Weight sharing mechanism in CNNs can drastically decrease the number of parameters. However, it may also restrict the models from learning other kinds of invariance. Tiled CNN [23] is a variation of CNN that tiles and multiples feature maps to learn rotational and scale invariant features. Separate kernels are learned within the same layer, and the complex invariances can be learned implicitly by square-root pooling over neighbouring units. As illustrated in Figure 3(b), the convolution operations are applied every $k$ unit, where $k$ is the tile size to control the distance over which weights are shared. When the tile size $k$ is 1, the units within each map will have the same weights, and tiled CNN becomes identical to the traditional CNN. In [23], their experiments on the NORB and CIFAR-10 datasets show that $k = 2$ achieves the best results. Wang *et al.* [24] find that Tiled CNN performs better than traditional CNN [25] on small time series datasets.

#### 3.1.2. Transposed Convolution

Transposed convolution can be seen as the backward pass of a corresponding traditional convolution. It is also known as deconvolution [11, 26–28] and fractionally strided convolution [29]. To stay consistent with most literature [11, 30], we use the term "deconvolution". Contrary to the traditional convolution that connects multiple input activations to a single activation, deconvolution associates a single activation with multiple output activations. Figure 3(d) shows a deconvolution operation of $3 \times 3$ kernel over a $4 \times 4$ input using unit stride and zero padding. The stride of deconvolution gives the dilation factor for the input feature map. Specifically, the deconvolution will first upsample the input by a factor of the stride value with padding, then perform convolution operation on the upsampled input. Recently, deconvolution has been widely used for visualization [11], recognition [31–33], localization [34], semantic segmentation [30], visual question answering [35], and super-resolution [36].

#### 3.1.3. Dilated Convolution

Dilated CNN [37] is a recent development of CNN that introduces one more hyper-parameter to the convolutional layer. By inserting zeros between filter elements, Dilated CNN can increase the network's receptive field size and let the network cover more relevant information. This is very important for tasks which need a large receptive field when making the prediction. Formally, a 1-D dilated convolution with dilation $l$ that convolves signal $F$ with kernel $k$ of size $r$ is defined as $(F *_l k)_t = \sum_\tau k_\tau F_{t-l\tau}$, where $*_l$ denotes $l$-dilated convolution. This formula can be straightforwardly extended to 2-D dilated convolution. Figure 3(c) shows an example of three dilated convolution layers where the dilation factor $l$ grows up exponentially at each layer. The middle feature map $F_2$ is produced from the bottom feature map $F_1$ by applying a 1-dilated convolution, where each element in $F_2$ has a receptive field of $3 \times 3$. $F_3$ is produced from $F_2$ by applying a 2-dilated convolution, where each element in $F_3$ has a receptive field of $(2^3 - 1) \times (2^3 - 1)$. The top feature map $F_4$ is produced from $F_3$ by applying a 4-dilated convolution, where each element in $F_4$ has a receptive field of $(2^4 - 1) \times (2^4 - 1)$. As can be seen, the size of receptive field of each element in $F_{i+1}$ is $(2^{i+2} - 1) \times (2^{(i+2)} - 1)$. Dilated CNNs have achieved impressive performance in tasks such as scene segmentation [37], machine translation [38], speech synthesis [39], and speech recognition [40].

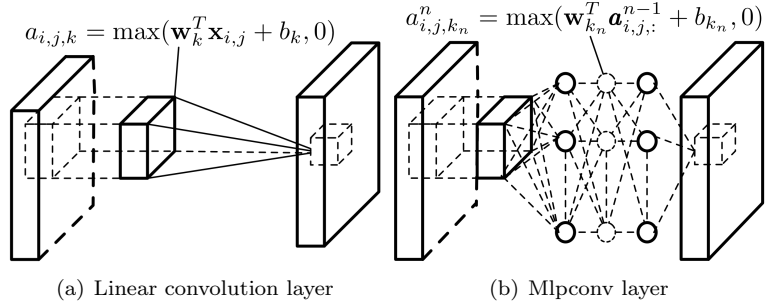

Figure 4: The comparison of linear convolution layer and mlpconv layer.

*3.1.4. Network in Network*

Network In Network (NIN) is a general network structure proposed by Lin *et al.* [18]. It replaces the linear filter of the convolutional layer by a micro network, *e.g.*, multilayer perceptron convolution (mlpconv) layer in the paper, which makes it capable of approximating more abstract representations of the latent concepts. The overall structure of NIN is the stacking of such micro networks. Figure 4 shows the difference between the linear convolutional layer and the mlpconv layer. Formally, the feature map of convolution layer (with nonlinear activation function, *e.g.*, ReLU [14]) is computed as:

$$a_{i,j,k} = \max(\mathbf{w}_k^T \mathbf{x}_{i,j} + b_k, 0) \qquad (5)$$

where $a_{i,j,k}$ is the activation value of $k$-th feature map at location $(i, j)$, $\mathbf{x}_{i,j}$ is the input patch centered at location $(i, j)$, $\mathbf{w}_k$ and $b_k$ are weight vector and bias term of the $k$-th filter. As a comparison, the computation performed by mlpconv layer is formulated as:

$$a_{i,j,k_n}^n = \max(\mathbf{w}_{k_n}^T \mathbf{a}_{i,j,:}^{n-1} + b_{k_n}, 0) \qquad (6)$$

where $n \in [1, N]$, $N$ is the number of layers in the mlpconv layer, $\mathbf{a}_{i,j,:}^0$ is equal to $\mathbf{x}_{i,j}$. In mlpconv layer, $1 \times 1$ convolutions are placed after the traditional convolutional layer. The $1 \times 1$ convolution is equivalent to the cross-channel parametric pooling operation which is succeeded by ReLU [14]. Therefore, the mlpconv layer can also be regarded as the cascaded cross-channel parametric pooling on the normal convolutional layer. In the end, they also apply a global average pooling which spatially averages the feature maps of the final layer, and directly feed the output vector into softmax layer. Compared with the fully-connected layer, global average pooling has fewer parameters and thus reduces the overfitting risk and computational load.

*3.1.5. Inception Module*

Inception module is introduced by Szegedy *et al.* [10] which can be seen as a logical culmination of NIN. They use variable filter sizes to capture different visual patterns of different sizes, and approximate the optimal sparse structure by the inception module. Specifically, inception module consists of one pooling operation and three types of convolution operations (see Figure 5(b)), and $1 \times 1$ convolutions are placed before $3 \times 3$ and $5 \times 5$ convolutions as dimension reduction modules, which allow for increasing the depth and width of CNN without increasing the computational complexity. With the help of inception module, the network parameters can be dramatically reduced to 5 millions which are much less than those of AlexNet (60 millions) and ZFNet (75 millions).

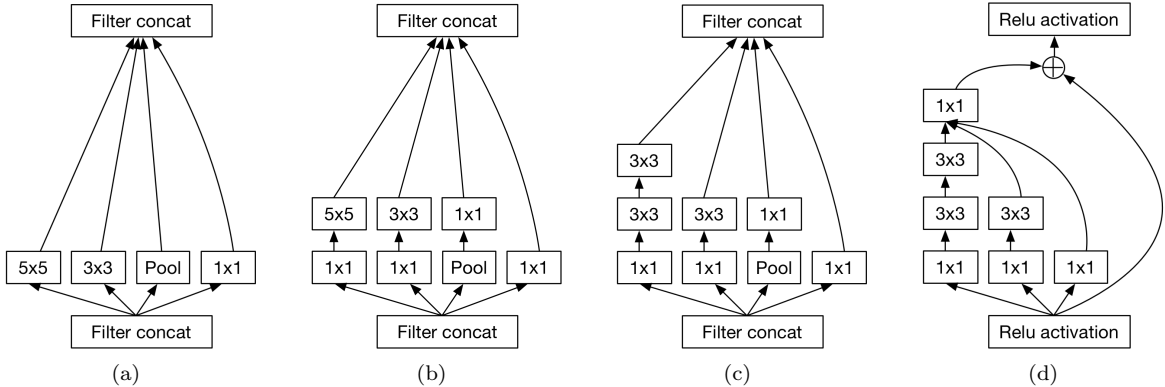

Figure 5: (a) Inception module, naive version. (b) The inception module used in [10]. (c) The improved inception module used in [41] where each $5 \times 5$ convolution is replaced by two $3 \times 3$ convolutions. (d) The Inception-ResNet-A module used in [42].

In their later paper [41], to find high-performance networks with a relatively modest computation cost, they suggest the representation size should gently decrease from inputs to outputs as well as spatial aggregation can be done over lower dimensional embeddings without much loss in representational power. The optimal performance of the network can be reached by balancing the number of filters per layer and the depth of the network. Inspired by the ResNet [12], their latest Inception-V4 [42] combines the inception architecture with shortcut connections (see Figure 5(d)). They find that shortcut connections can significantly accelerate the training of inception networks. Their Inception-v4 model architecture (with 75 trainable layers) that ensembles three residual and one Inception-v4 can achieve 3.08% top-5 error rate on the validation dataset of ILSVRC 2012.

*3.2. Pooling Layer*

Pooling is an important concept of CNN. It lowers the computational burden by reducing the number of connections between convolutional layers. In this section, we introduce some recent pooling methods used in CNNs.

*3.2.1. $L_p$ Pooling*

$L_p$ pooling is a biologically inspired pooling process modelled on complex cells [43]. It has been theoretically analyzed in [44], which suggest that $L_p$ pooling provides better generalization than max pooling. $L_p$ pooling can be represented as:

$$y_{i,j,k} = [\sum_{(m,n)\in \mathcal{R}_{ij}} (a_{m,n,k})^p]^{1/p} \tag{7}$$

where $y_{i,j,k}$ is the output of the pooling operator at location $(i, j)$ in $k$-th feature map, and $a_{m,n,k}$ is the feature value at location $(m, n)$ within the pooling region $\mathcal{R}_{ij}$ in $k$-th feature map. Specially, when $p = 1$, $L_p$ corresponds to average pooling, and when $p = \infty$, $L_p$ reduces to max pooling.

*3.2.2. Mixed Pooling*

Inspired by random Dropout [17] and DropConnect [45], Yu et al. [46] propose a mixed pooling method which is the combination of max pooling and average pooling. The function of mixed pooling can be

formulated as follows:

$$y_{i,j,k} = \lambda \max_{(m,n) \in \mathcal{R}_{ij}} a_{m,n,k} + (1-\lambda) \frac{1}{|\mathcal{R}_{ij}|} \sum_{(m,n) \in \mathcal{R}_{ij}} a_{m,n,k} \qquad (8)$$

where $\lambda$ is a random value being either 0 or 1 which indicates the choice of either using average pooling or max pooling. During forward propagation process, $\lambda$ is recorded and will be used for the backpropagation operation. Experiments in [46] show that mixed pooling can better address the overfitting problems and it performs better than max pooling and average pooling.

*3.2.3. Stochastic Pooling*

Stochastic pooling [47] is a dropout-inspired pooling method. Instead of picking the maximum value within each pooling region as max pooling does, stochastic pooling randomly picks the activations according to a multinomial distribution, which ensures that the non-maximal activations of feature maps are also possible to be utilized. Specifically, stochastic pooling first computes the probabilities $p$ for each region $R_j$ by normalizing the activations within the region, *i.e.*, $p_i = a_i / \sum_{k \in \mathcal{R}_j}(a_k)$. After obtaining the distribution $P(p_1, ..., p_{|\mathcal{R}_j|})$, we can sample from the multinomial distribution based on $p$ to pick a location $l$ within the region, and then set the pooled activation as $y_j = a_l$, where $l \sim P(p_1, ..., p_{|\mathcal{R}_j|})$. Compared with max pooling, stochastic pooling can avoid overfitting due to the stochastic component.

*3.2.4. Spectral Pooling*

Spectral pooling [48] performs dimensionality reduction by cropping the representation of input in frequency domain. Given an input feature map $\mathbf{x} \in \mathbb{R}^{m \times m}$, suppose the dimension of desired output feature map is $h \times w$, spectral pooling first computes the discrete Fourier transform (DFT) of the input feature map, then crops the frequency representation by maintaining only the central $h \times w$ submatrix of the frequencies, and finally uses inverse DFT to map the approximation back into spatial domain. Compared with max pooling, the linear low-pass filtering operation of spectral pooling can preserve more information for the same output dimensionality. Meanwhile, it also does not suffer from the sharp reduction in output map dimensionality exhibited by other pooling methods. What is more, the process of spectral pooling is achieved by matrix truncation, which makes it capable of being implemented with little computational cost in CNNs (*e.g.*, [49]) that employ FFT for convolution kernels.

*3.2.5. Spatial Pyramid Pooling*

Spatial pyramid pooling (SPP) is introduced by He *et al.* [50]. The key advantage of SPP is that it can generate a fixed-length representation regardless of the input sizes. SPP pools input feature map in local spatial bins with sizes proportional to the image size, resulting in a fixed number of bins. This is different from the sliding window pooling in the previous deep networks, where the number of sliding windows depends on the input size. By replacing the last pooling layer with SPP, they propose a new SPP-net which is able to deal with images with different sizes.

*3.2.6. Multi-scale Orderless Pooling*

Inspired by [51], Gong *et al.* [52] use multi-scale orderless pooling (MOP) to improve the invariance of CNNs without degrading their discriminative power. They extract deep activation features for both the whole image and local patches of several scales. The activations of the whole image are the same as those of previous CNNs, which aim to capture the global spatial layout information. The activations of local patches are aggregated by VLAD encoding [53], which aim to capture more local, fine-grained details of the image as well as enhancing invariance. The new image representation is obtained by concatenating the global activations and the VLAD features of the local patch activations.

*3.3. Activation Function*

A proper activation function significantly improves the performance of a CNN for a certain task. In this section, we introduce the recently used activation functions in CNNs.

*3.3.1. ReLU*

Rectified linear unit (ReLU) [14] is one of the most notable non-saturated activation functions. The ReLU activation function is defined as:

$$a_{i,j,k} = \max(z_{i,j,k}, 0) \tag{9}$$

where $z_{i,j,k}$ is the input of the activation function at location $(i,j)$ on the $k$-th channel. ReLU is a piecewise linear function which prunes the negative part to zero and retains the positive part (see Figure 6(a)). The simple $\max(\cdot)$ operation of ReLU allows it to compute much faster than sigmoid or tanh activation functions, and it also induces the sparsity in the hidden units and allows the network to easily obtain sparse representations. It has been shown that deep networks can be trained efficiently using ReLU even without pre-training [8]. Even though the discontinuity of ReLU at 0 may hurt the performance of backpropagation, many works have shown that ReLU works better than sigmoid and tanh activation functions empirically [54, 55].

*3.3.2. Leaky ReLU*

A potential disadvantage of ReLU unit is that it has zero gradient whenever the unit is not active. This may cause units that do not active initially never active as the gradient-based optimization will not adjust their weights. Also, it may slow down the training process due to the constant zero gradients. To alleviate this problem, Mass *et al.* introduce Leaky ReLU (LReLU) [54] which is defined as:

$$a_{i,j,k} = \max(z_{i,j,k}, 0) + \lambda \min(z_{i,j,k}, 0) \tag{10}$$

where $\lambda$ is a predefined parameter in range $(0, 1)$. Compared with ReLU, Leaky ReLU compresses the negative part rather than mapping it to constant zero, which makes it allow for a small, non-zero gradient when the unit is not active.

*3.3.3. Parametric ReLU*

Rather than using a predefined parameter in Leaky ReLU, *e.g.*, $\lambda$ in Eq.(10), He *et al.* [56] propose Parametric Rectified Linear Unit (PReLU) which adaptively learns the parameters of the rectifiers in order to improve accuracy. Mathematically, PReLU function is defined as:

$$a_{i,j,k} = \max(z_{i,j,k}, 0) + \lambda_k \min(z_{i,j,k}, 0) \tag{11}$$

where $\lambda_k$ is the learned parameter for the $k$-th channel. As PReLU only introduces a very small number of extra parameters, *e.g.*, the number of extra parameters is the same as the number of channels of the whole network, there is no extra risk of overfitting and the extra computational cost is negligible. It also can be simultaneously trained with other parameters by backpropagation.

*3.3.4. Randomized ReLU*

Another variant of Leaky ReLU is Randomized Leaky Rectified Linear Unit (RReLU) [57]. In RReLU, the parameters of negative parts are randomly sampled from a uniform distribution in training, and then fixed in testing (see Figure 6(c)). Formally, RReLU function is defined as:

$$a_{i,j,k}^{(n)} = \max(z_{i,j,k}^{(n)}, 0) + \lambda_k^{(n)} \min(z_{i,j,k}^{(n)}, 0) \tag{12}$$

where $z_{i,j,k}^{(n)}$ denotes the input of activation function at location $(i, j)$ on the $k$-th channel of $n$-th example, $\lambda_k^{(n)}$ denotes its corresponding sampled parameter, and $a_{i,j,k}^{(n)}$ denotes its corresponding output. It could reduce overfitting due to its randomized nature. Xu *et al.* [57] also evaluate ReLU, LReLU, PReLU and RReLU on standard image classification task, and concludes that incorporating a non-zero slop for negative part in rectified activation units could consistently improve the performance.

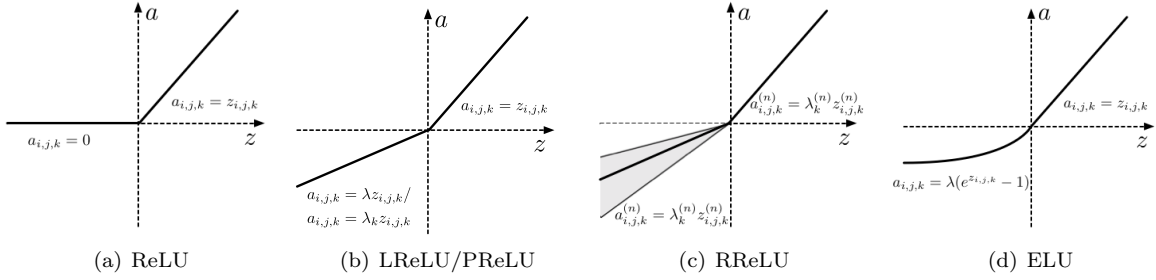

Figure 6: The comparison among ReLU, LReLU, PReLU, RReLU and ELU. For Leaky ReLU, $\lambda$ is empirically predefined. For PReLU, $\lambda_k$ is learned from training data. For RReLU, $\lambda_k^{(n)}$ is a random variable which is sampled from a given uniform distribution in training and keeps fixed in testing. For ELU, $\lambda$ is empirically predefined.

*3.3.5. ELU*

Clevert *et al.* [58] introduce Exponential Linear Unit (ELU) which enables faster learning of deep neural networks and leads to higher classification accuracies. Like ReLU, LReLU, PReLU and RReLU, ELU avoids the vanishing gradient problem by setting the positive part to identity. In contrast to ReLU, ELU has a negative part which is beneficial for fast learning.

Compared with LReLU, PReLU, and RReLU which also have unsaturated negative parts, ELU employs a saturation function as negative part. As the saturation function will decrease the variation of the units if deactivated, it makes ELU more robust to noise. The function of ELU is defined as:

$$a_{i,j,k} = \max(z_{i,j,k}, 0) + \min(\lambda(e^{z_{i,j,k}} - 1), 0) \tag{13}$$

where $\lambda$ is a predefined parameter for controlling the value to which an ELU saturate for negative inputs.

*3.3.6. Maxout*

Maxout [59] is an alternative nonlinear function that takes the maximum response across multiple channels at each spatial position. As stated in [59], the maxout function is defined as: $a_{i,j,k} = \max_{k \in [1,K]} z_{i,j,k}$, where $z_{i,j,k}$ is the $k$-th channel of the feature map. It is worth noting that maxout enjoys all the benefits of ReLU since ReLU is actually a special case of maxout, *e.g.*, $\max(\mathbf{w}_1^T \mathbf{x} + b_1, \mathbf{w}_2^T \mathbf{x} + b_2)$ where $\mathbf{w}_1$ is a zero vector and $b_1$ is zero. Besides, maxout is particularly well suited for training with Dropout.

*3.3.7. Probout*

Springenberg *et al.* [60] propose a probabilistic variant of maxout called probout. They replace the maximum operation in maxout with a probabilistic sampling procedure. Specifically, they first define a probability for each of the $k$ linear units as: $p_i = e^{\lambda z_i} / \sum_{j=1}^{k} e^{\lambda z_j}$, where $\lambda$ is a hyperparameter for controlling the variance of the distribution. Then, they pick one of the k units according to a multinomial distribution $\{p_1, ..., p_k\}$ and set the activation value to be the value of the picked unit. In order to incorporate with dropout, they actually re-define the probabilities as:

$$\hat{p}_0 = 0.5, \quad \hat{p}_i = e^{\lambda z_i} / (2 \cdot \sum_{j=1}^{k} e^{\lambda z_j}) \tag{14}$$

The activation function is then sampled as:

$$a_i = \begin{cases} 0 & \text{if } i = 0 \\ z_i & \text{else} \end{cases} \tag{15}$$

where $i \sim \text{multinomial}\{\hat{p}_0, ..., \hat{p}_k\}$. Probout can achieve the balance between preserving the desirable properties of maxout units and improving their invariance properties. However, in testing process, probout is computationally expensive than maxout due to the additional probability calculations.

### 3.4. Loss Function

It is important to choose an appropriate loss function for a specific task. We introduce four representative ones in this subsection: Hinge loss, Softmax loss, Contrastive loss, Triplet loss.

#### 3.4.1. Hinge Loss

Hinge loss is usually used to train large margin classifiers such as Support Vector Machine (SVM). The hinge loss function of a multi-class SVM is defined in Eq.(16), where $\mathbf{w}$ is the weight vector of classifier and $\boldsymbol{y}^{(i)} \in [1, \ldots, K]$ indicates its correct class label among the $K$ classes.

$$\mathcal{L}_{hinge} = \frac{1}{N} \sum_{i=1}^{N} \sum_{j=1}^{K} [\max(0, 1 - \delta(\boldsymbol{y}^{(i)}, j)\mathbf{w}^T \mathbf{x}_i)]^p \tag{16}$$

where $\delta(\boldsymbol{y}^{(i)}, j) = 1$ if $\boldsymbol{y}^{(i)} = j$, otherwise $\delta(\boldsymbol{y}^{(i)}, j) = -1$. Note that if $p = 1$, Eq.(16) is Hinge-Loss ($L_1$-Loss), while if $p = 2$, it is the Squared Hinge-Loss ($L_2$-Loss) [61]. The $L_2$-Loss is differentiable and imposes a larger loss for point which violates the margin comparing with $L_1$-Loss. [19] investigates and compares the performance of softmax with $L_2$-SVMs in deep networks. The results on MNIST [62] demonstrate the superiority of $L_2$-SVM over softmax.

#### 3.4.2. Softmax Loss

Softmax loss is a commonly used loss function which is essentially a combination of multinomial logistic loss and softmax. Given a training set $\{(\boldsymbol{x}^{(i)}, \boldsymbol{y}^{(i)}); i \in 1, \ldots, N, \boldsymbol{y}^{(i)} \in 1, \ldots, K\}$, where $\boldsymbol{x}^{(i)}$ is the $i$-th input image patch, and $\boldsymbol{y}^{(i)}$ is its target class label among the $K$ classes. The prediction of $j$-th class for $i$-th input is transformed with the softmax function: $p_j^{(i)} = e^{z_j^{(i)}} / \sum_{l=1}^{K} e^{z_l^{(i)}}$, where $z_j^{(i)}$ is usually the activations of a densely connected layer, so $z_j^{(i)}$ can be written as $z_j^{(i)} = \mathbf{w}_j^T \mathbf{a}^{(i)} + \mathbf{b}_j$. Softmax turns the predictions into non-negative values and normalizes them to get a probability distribution over classes. Such probabilistic predictions are used to compute the multinomial logistic loss, *i.e.*, the softmax loss, as follows:

$$\mathcal{L}_{softmax} = -\frac{1}{N} [\sum_{i=1}^{N} \sum_{j=1}^{K} 1\{\boldsymbol{y}^{(i)} = j\} \log p_j^{(i)}] \tag{17}$$

Recently, Liu *et al.* [63] propose the Large-Margin Softmax (L-Softmax) loss, which introduces an angular margin to the angle $\theta_j$ between input feature vector $\mathbf{a}^{(i)}$ and the $j$-th column $\mathbf{w}_j$ of weight matrix. The prediction $p_j^{(i)}$ for L-Softmax loss is defined as:

$$p_j^{(i)} = \frac{e^{\|\mathbf{w}_j\| \|\mathbf{a}^{(i)}\| \psi(\theta_j)}}{e^{\|\mathbf{w}_j\| \|\mathbf{a}^{(i)}\| \psi(\theta_j)} + \sum_{l \neq j} e^{\|\mathbf{w}_l\| \|\mathbf{a}^{(i)}\| \cos(\theta_l)}} \tag{18}$$

$$\psi(\theta_j) = (-1)^k \cos(m\theta_j) - 2k, \theta_j \in [k\pi/m, (k+1)\pi/m] \tag{19}$$

where $k \in [0, m-1]$ is an integer, $m$ controls the margin among classes. When $m = 1$, the L-Softmax loss reduces to the original softmax loss. By adjusting the margin $m$ between classes, a relatively difficult learning objective will be defined, which can effectively avoid overfitting. They verify the effective of L-Softmax on MNIST, CIFAR-10, and CIFAR-100, and find that the L-Softmax loss performs better than the original softmax.

#### 3.4.3. Contrastive Loss

Contrastive loss is commonly used to train Siamese network [64–67] which is a weakly-supervised scheme for learning a similarity measure from pairs of data instances labelled as matching or non-matching. Given the $i$-th pair of data $(\boldsymbol{x}_\alpha^{(i)}, \boldsymbol{x}_\beta^{(i)})$, let $(\boldsymbol{z}_\alpha^{(i,l)}, \boldsymbol{z}_\beta^{(i,l)})$ denotes its corresponding output pair of the $l$-th ($l \in [1, \cdots, L]$) layer. In [65] and [66], they pass the image pairs through two identical CNNs, and feed the feature vectors of the final layer to the cost module. The contrastive loss function that they use for training samples is:

$$\mathcal{L}_{contrastive} = \frac{1}{2N} \sum_{i=1}^{N} (y)d^{(i,L)} + (1-y)\max(m - d^{(i,L)}, 0) \qquad (20)$$

where $d^{(i,L)} = ||z_\alpha^{(i,L)} - z_\beta^{(i,L)}||_2^2$, and $m$ is a margin parameter affecting non-matching pairs. If $(x_\alpha^{(i)}, x_\beta^{(i)})$ is a matching pair, then y = 1. Otherwise, y = 0.

Lin et al. [68] find that such a single margin loss function causes a dramatic drop in retrieval results when fine-tuning the network on all pairs. Meanwhile, the performance is better retained when fine-tuning only on non-matching pairs. This indicates that the handling of matching pairs in the loss function is responsible for the drop. While the recall rate on non-matching pairs alone is stable, handling the matching pairs is the main reason for the drop in recall rate. To solve this problem, they propose a double margin loss function which adds another margin parameter to affect the matching pairs. Instead of calculating the loss of the final layer, their contrastive loss is defined for every layer $l$ and the backpropagations for the loss of individual layers are performed at the same time. It is defined as:

$$\mathcal{L}_{d-contrastive} = \frac{1}{2N} \sum_{i=1}^{N} \sum_{l=1}^{L} (y)\max(d^{(i,l)} - m_1, 0) + (1-y)\max(m_2 - d^{(i,l)}, 0) \qquad (21)$$

In practice, they find that these two margin parameters can set to be equal ($m_1 = m_2 = m$) and be learned from the distribution of the sampled matching and non-matching image pairs.

### 3.4.4. Triplet Loss

Triplet loss [69] considers three instances per loss function. The triplet units $(x_a^{(i)}, x_p^{(i)}, x_n^{(i)})$ usually contain an anchor instance $x_a^{(i)}$ as well as a positive instance $x_p^{(i)}$ from the same class of $x_a^{(i)}$ and a negative instance $x_n^{(i)}$. Let $(z_a^{(i)}, z_p^{(i)}, z_n^{(i)})$ denote the feature representation of the triplet units, the triplet loss is defined as:

$$\mathcal{L}_{triplet} = \frac{1}{N} \sum_{i=1}^{N} \max\{d_{(a,p)}^{(i)} - d_{(a,n)}^{(i)} + m, 0\} \qquad (22)$$

where $d_{(a,p)}^{(i)} = ||z_a^{(i)} - z_p^{(i)}||_2^2$ and $d_{(a,n)}^{(i)} = ||z_a^{(i)} - z_n^{(i)}||_2^2$. The objective of triplet loss is to minimize the distance between the anchor and positive, and maximize the distance between the negative and the anchor.

However, randomly selected anchor samples may judge falsely in some special cases. For example, when $d_{(n,p)}^{(i)} < d_{(a,p)}^{(i)} < d_{(a,n)}^{(i)}$, the triplet loss may still be zero. Thus the triplet units will be neglected during the backward propagation. Liu et al. [70] propose the Coupled Clusters (CC) loss to solve this problem. Instead of using the triplet units, the coupled clusters loss function is defined over the positive set and the negative set. By replacing the randomly picked anchor with the cluster center, it makes the samples in the positive set cluster together and samples in the negative set stay relatively far away, which is more reliable than the original triplet loss. The coupled clusters loss function is defined as:

$$\mathcal{L}_{cc} = \frac{1}{N^p} \sum_{i=1}^{N^p} \frac{1}{2} \max\{||z_p^{(i)} - c_p||_2^2 - ||z_n^{(*)} - c_p||_2^2 + m, 0\} \qquad (23)$$

where $N^p$ is the number of samples per set, $z_n^{(*)}$ is the feature representation of $x_n^{(*)}$ which is the nearest negative sample to the estimated center point $c_p = (\sum_i^{N^p} z_p^{(i)})/N^p$. Triplet loss and its variants have been widely used in various tasks, including re-identification [71], verification [70], and image retrieval [72].

### 3.4.5. Kullback-Leibler Divergence

Kullback-Leibler Divergence (KLD) is a non-symmetric measure of the difference between two probability distributions $p(x)$ and $q(x)$ over the same discrete variable $x$ (see Figure 7(a)). The KLD from $q(x)$ to $p(x)$

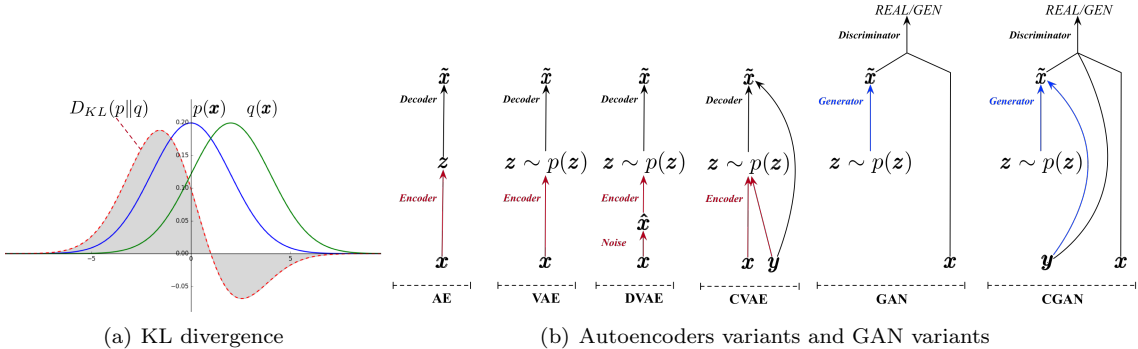

(a) KL divergence      (b) Autoencoders variants and GAN variants

Figure 7: The illustration of (a) the KullbackLeibler divergence for two normal Gaussian distributions, (b) AE variants (AE, VAE [73], DVAE [74], and CVAE [75]) and GAN variants (GAN [76], CGAN [77]).

is defined as:

$$\mathcal{D}_{KL}(p||q) = -H(p(x)) - E_p[\log q(x)] \tag{24}$$

$$= \sum_x p(x) \log p(x) - \sum_x p(x) \log q(x) = \sum_x p(x) \log \frac{p(x)}{q(x)} \tag{25}$$

where $H(p(x))$ is the Shannon entropy of $p(x)$, $E_p(\log q(x))$ is the cross entropy between $p(x)$ and $q(x)$.

KLD has been widely used as a measure of information loss in the objective function of various Autoencoders (AEs) [78–80]. Famous variants of AE include sparse AE [81, 82], Denoising AE [78] and Variational AE (VAE) [73]. VAE interprets the latent representation through Bayesian inference. It consists of two parts: an encoder which "compresses" the data sample $x$ to the latent representation $z \sim q_\phi(z|x)$; and a decoder, which maps such representation back to data space $\tilde{x} \sim p_\theta(x|z)$ which as close to the input as possible, where $\phi$ and $\theta$ are the parameters of encoder and decoder respectively. As proposed in [73], VAEs try to maximize the variational lower bound of the log-likelihood of $\log p(x|\phi, \theta)$:

$$\mathcal{L}_{vae} = \mathrm{E}_{z \sim q_\phi(z|x)}[\log p_\theta(x|z)] - \mathcal{D}_{KL}(q_\phi(z|x) \| p(z)) \tag{26}$$

where the first term is the reconstruction cost, and the KLD term enforces prior $p(z)$ on the proposal distribution $q_\phi(z|x)$. Usually $p(z)$ is the standard normal distribution [73], discrete distribution [75], or some distributions with geometric interpretation [83]. Following the original VAE, many variants have been proposed [74, 75, 84]. Conditional VAE (CVAE) [75, 84] generates samples from the conditional distribution with $\tilde{x} \sim p_\theta(x|y, z)$. Denoising VAE (DVAE) [74] recovers the original input $x$ from the corrupted input $\hat{x}$ [78].

Jensen-Shannon Divergence (JSD) is a symmetrical form of KLD. It measures the similarity between $p(x)$ and $q(x)$:

$$\mathcal{D}_{JS}(p||q) = \frac{1}{2}\mathcal{D}_{KL}\left(p(x) \left\| \frac{p(x) + q(x)}{2}\right.\right) + \frac{1}{2}\mathcal{D}_{KL}\left(q(x) \left\| \frac{p(x) + q(x)}{2}\right.\right) \tag{27}$$

By minimizing the JSD, we can make the two distributions $p(x)$ and $q(x)$ as close as possible. JSD has been successfully used in the Generative Adversarial Networks (GANs) [76, 85, 86]. In contrast to VAEs that model the relationship between $x$ and $z$ directly, GANs are explicitly set up to optimize for generative tasks [85]. The objective of GANs is to find the discriminator $D$ that gives the best discrimination between the real and generated data, and simultaneously encourage the generator $G$ to fit the real data distribution. The min-max game played between the discriminator $D$ and the generator $G$ is formalized by the following objective function:

$$\min_G \max_D \mathcal{L}_{gan}(D, G) = \mathbb{E}_{x \sim p(x)}[\log D(x)] + \mathbb{E}_{z \sim q(z)}[\log(1 - D(G(z)))] \tag{28}$$

The original GAN paper [76] shows that for a fixed generator $G^*$, we have the optimal discriminator $D_G^*(x) = \frac{p(x)}{p(x)+q(x)}$. Then the Equation 28 is equivalent to minimize the JSD between $p(x)$ and $q(x)$. If $G$ and $D$ have enough capacity, the distribution $q(x)$ converges to $p(x)$. Like Conditional VAE, the Conditional GAN (CGAN) [77] also receives an additional information $y$ as input to generate samples conditioning on $y$. In practice, GANs are notoriously unstable to train [87, 88].

### 3.5. Regularization

Overfitting is an unneglectable problem in deep CNNs, which can be effectively reduced by regularization. In the following subsection, we introduce some effective regularization techniques: $\ell_p$-norm, Dropout, and DropConnect.

#### 3.5.1. $\ell_p$-norm Regularization

Regularization modifies the objective function by adding additional terms that penalize the model complexity. Formally, if the loss function is $\mathcal{L}(\theta, \mathbf{x}, \mathbf{y})$, then the regularized loss will be:

$$E(\theta, \mathbf{x}, \mathbf{y}) = \mathcal{L}(\theta, \mathbf{x}, \mathbf{y}) + \lambda R(\theta) \tag{29}$$

where $R(\theta)$ is the regularization term, and $\lambda$ is the regularization strength.

$\ell_p$-norm regularization function is usually employed as $R(\theta) = \sum_j \|\theta_j\|_p^p$. When $p \geq 1$, the $\ell_p$-norm is convex, which makes the optimization easier and renders this function attractive [17]. For $p = 2$, the $\ell_2$-norm regularization is commonly referred to as weight decay. A more principled alternative of $\ell_2$-norm regularization is Tikhonov regularization [89], which rewards invariance to noise in the inputs. When $p < 1$, the $\ell_p$-norm regularization more exploits the sparsity effect of the weights but conducts to non-convex function.

#### 3.5.2. Dropout

Dropout is first introduced by Hinton *et al.* [17], and it has been proven to be very effective in reducing overfitting. In [17], they apply Dropout to fully-connected layers. The output of Dropout is $\mathbf{y} = \mathbf{r} * a(\mathbf{W}^T \mathbf{x})$, where $\mathbf{x} = [x_1, x_2, \ldots, x_n]^T$ is the input to fully-connected layer, $\mathbf{W} \in \mathbb{R}^{n \times d}$ is a weight matrix, and $\mathbf{r}$ is a binary vector of size $d$ whose elements are independently drawn from a Bernoulli distribution with parameter $p$, i.e. $\mathrm{r}_i \sim Bernoulli(p)$. Dropout can prevent the network from becoming too dependent on any one (or any small combination) of neurons, and can force the network to be accurate even in the absence of certain information. Several methods have been proposed to improve Dropout. Wang *et al.* [90] propose a fast Dropout method which can perform fast Dropout training by sampling from or integrating a Gaussian approximation. Ba *et al.* [91] propose an adaptive Dropout method, where the Dropout probability for each hidden variable is computed using a binary belief network that shares parameters with the deep network. In [92], they find that applying standard Dropout before $1 \times 1$ convolutional layer generally increases training time but does not prevent overfitting. Therefore, they propose a new Dropout method called SpatialDropout, which extends the Dropout value across the entire feature map. This new Dropout method works well especially when the training data size is small.

#### 3.5.3. DropConnect

DropConnect [45] takes the idea of Dropout a step further. Instead of randomly setting the outputs of neurons to zero, DropConnect randomly sets the elements of weight matrix $\mathbf{W}$ to zero. The output of DropConnect is given by $\mathbf{y} = \mathrm{a}((\mathbf{R} * \mathbf{W})\mathbf{x})$, where $\mathrm{R}_{ij} \sim Bernoulli(p)$. Additionally, the biases are also masked out during the training process. Figure 8 illustrates the differences among No-Drop, Dropout and DropConnect networks.

### 3.6. Optimization

In this subsection, we discuss some key techniques for optimizing CNNs.

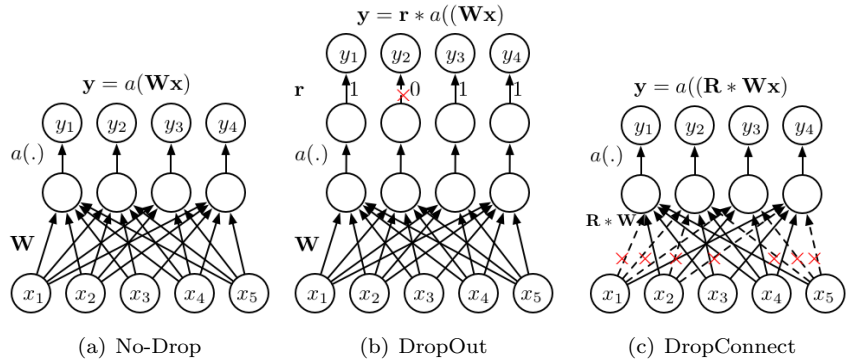

Figure 8: The illustration of No-Drop network, DropOut network and DropConnect network.

*3.6.1. Data Augmentation*

Deep CNNs are particularly dependent on the availability of large quantities of training data. An elegant solution to alleviate the relative scarcity of the data compared to the number of parameters involved in CNNs is data augmentation [8]. Data augmentation consists in transforming the available data into new data without altering their natures. Popular augmentation methods include simple geometric transformations such as sampling [8], mirroring [93], rotating [94], shifting [95], and various photometric transformations [96]. Paulin *et al.* [97] propose a greedy strategy that selects the best transformation from a set of candidate transformations. However, their strategy involves a large number of model re-training steps, which can be computationally expensive when the number of candidate transformations is large. Hauberg *et al.* [98] propose an elegant way for data augmentation by randomly generating diffeomorphisms. Xie *et al.* [99] and Xu *et al.* [100] offer additional means of collecting images from the Internet to improve learning in fine-grained recognition tasks.

*3.6.2. Weight Initialization*

Deep CNN has a huge amount of parameters and its loss function is non-convex [101], which makes it very difficult to train. To achieve a fast convergence in training and avoid the vanishing gradient problem, a proper network initialization is one of the most important prerequisites [102, 103]. The bias parameters can be initialized to zero, while the weight parameters should be initialized carefully to break the symmetry among hidden units of the same layer. If the network is not properly initialized, *e.g.*, each layer scales its input by $k$, the final output will scale the original input by $k^L$ where $L$ is the number of layers. In this case, the value of $k > 1$ leads to extremely large values of output layers while the value of $k < 1$ leads a diminishing output value and gradients. Krizhevsky *et al.* [8] initialize the weights of their network from a zero-mean Gaussian distribution with standard deviation 0.01 and set the bias terms of the second, fourth and fifth convolutional layers as well as all the fully-connected layers to constant one. Another famous random initialization method is "Xavier", which is proposed in [104]. They pick the weights from a Gaussian distribution with zero mean and a variance of $2/(n_{\text{in}} + n_{\text{out}})$, where $n_{\text{in}}$ is the number of neurons feeding into it, and $n_{\text{out}}$ is the number of neurons the result is fed to. Thus "Xavier" can automatically determine the scale of initialization based on the number of input and output neurons, and keep the signal in a reasonable range of values through many layers. One of its variants in Caffe [1] uses the $n_{\text{in}}$-only variant, which makes it much easier to implement. "Xavier" initialization method is later extended by [56] to account for the rectifying nonlinearities, where they derive a robust initialization method that particularly considers the ReLU nonlinearity. Their method, allows for the training of extremely deep models (*e.g.*, [10]) to converge while the "Xavier" method [104] cannot.

---

[1] https://github.com/BVLC/caffe

Independently, Saxe *et al.* [105] show that orthonormal matrix initialization works much better for linear networks than Gaussian initialization, and it also works for networks with nonlinearities. Mishkin *et al.* [102] extend [105] to an iterative procedure. Specifically, it proposes a layer-sequential unit-variance process scheme which can be viewed as an orthonormal initialization combined with batch normalization (see Section 3.6.4) performed only on the first mini-batch. It is similar to batch normalization as both of them take a unit variance normalization procedure. Differently, it uses ortho-normalization to initialize the weights which helps to efficiently de-correlate layer activities. Such an initialization technique has been applied to [106, 107] with a remarkable increase in performance.

*3.6.3. Stochastic Gradient Descent*

The backpropagation algorithm is the standard training method which uses gradient descent to update the parameters. Many gradient descent optimization algorithms have been proposed [108, 109]. Standard gradient descent algorithm updates the parameters $\boldsymbol{\theta}$ of the objective $\mathcal{L}(\boldsymbol{\theta})$ as $\boldsymbol{\theta}_{t+1} = \boldsymbol{\theta}_t - \eta \nabla_{\boldsymbol{\theta}} E[\mathcal{L}(\boldsymbol{\theta}_t)]$, where $E[\mathcal{L}(\boldsymbol{\theta}_t)]$ is the expectation of $\mathcal{L}(\boldsymbol{\theta})$ over the full training set and $\eta$ is the learning rate. Instead of computing $E[\mathcal{L}(\boldsymbol{\theta}_t)]$, Stochastic Gradient Descent (SGD) [21] estimates the gradients on the basis of a single randomly picked example $(\boldsymbol{x}^{(t)}, \boldsymbol{y}^{(t)})$ from the training set:

$$\boldsymbol{\theta}_{t+1} = \boldsymbol{\theta}_t - \eta_t \nabla_{\boldsymbol{\theta}} \mathcal{L}(\boldsymbol{\theta}_t; \boldsymbol{x}^{(t)}, \boldsymbol{y}^{(t)}) \tag{30}$$

In practice, each parameter update in SGD is computed with respect to a mini-batch as opposed to a single example. This could help to reduce the variance in the parameter update and can lead to more stable convergence. The convergence speed is controlled by the learning rate $\eta_t$. However, mini-batch SGD does not guarantee good convergence, and there are still some challenges that need to be addressed. Firstly, it is not easy to choose a proper learning rate. One common method is to use a constant learning rate that gives stable convergence in the initial stage, and then reduce the learning rate as the convergence slows down. Additionally, learning rate schedules [110, 111] have been proposed to adjust the learning rate during the training. To make the current gradient update depend on historical batches and accelerate training, momentum [108] is proposed to accumulate a velocity vector in the relevant direction. The classical momentum update is given by:

$$\boldsymbol{v}_{t+1} = \gamma \boldsymbol{v}_t - \eta_t \nabla_{\boldsymbol{\theta}} \mathcal{L}(\boldsymbol{\theta}_t; \boldsymbol{x}^{(t)}, \boldsymbol{y}^{(t)}) \tag{31}$$

$$\boldsymbol{\theta}_{t+1} = \boldsymbol{\theta}_t + \boldsymbol{v}_{t+1} \tag{32}$$

where $\boldsymbol{v}_{t+1}$ is the current velocity vector, $\gamma$ is the momentum term which is usually set to 0.9. Nesterov momentum [103] is another way of using momentum in gradient descent optimization:

$$\boldsymbol{v}_{t+1} = \gamma \boldsymbol{v}_t - \eta_t \nabla_{\boldsymbol{\theta}} \mathcal{L}(\boldsymbol{\theta}_t + \gamma \boldsymbol{v}_t; \boldsymbol{x}^{(t)}, \boldsymbol{y}^{(t)}) \tag{33}$$

Compared with the classical momentum [108] which first computes the current gradient and then moves in the direction of the updated accumulated gradient, Nesterov momentum first moves in the direction of the previous accumulated gradient $\gamma \boldsymbol{v}_t$, calculates the gradient and then makes a gradient update. This anticipatory update prevents the optimization from moving too fast and achieves better performance [112].

Parallelized SGD methods [22, 113] improve SGD to be suitable for parallel, large-scale machine learning. Unlike standard (synchronous) SGD in which the training will be delayed if one of the machines is slow, these parallelized methods use the asynchronous mechanism so that no other optimizations will be delayed except for the one on the slowest machine. Jeffrey Dean *et al.* [114] use another asynchronous SGD procedure called Downpour SGD to speed up the large-scale distributed training process on clusters with many CPUs. There are also some works that use asynchronous SGD with multiple GPUs. Paine *et al.* [115] basically combine asynchronous SGD with GPUs to accelerate the training time by several times compared to training on a single machine. Zhuang *et al.* [116] also use multiple GPUs to asynchronously calculate gradients and update the global model parameters, which achieves 3.2 times of speedup on 4 GPUs compared to training on a single GPU.

Note that SGD methods may not result in convergence. The training process can be terminated when the performance stops improving. A popular remedy to over-training is to use early stopping [117] in which optimization is halted based on the performance on a validation set during training. To control the duration of the training process, various stopping criteria can be considered. For example, the training might be performed for a fixed number of epochs, or until a predefined training error is reached [118]. The stopping strategy should be done carefully [119], a proper stopping strategy should let the training process continue as long as the network generalization ability is improved and the overfitting is avoided.

*3.6.4. Batch Normalization*

Data normalization is usually the first step of data preprocessing. Global data normalization transforms all the data to have zero-mean and unit variance. However, as the data flows through a deep network, the distribution of input to internal layers will be changed, which will lose the learning capacity and accuracy of the network. Ioffe *et al.* [120] propose an efficient method called Batch Normalization (BN) to partially alleviate this phenomenon. It accomplishes the so-called covariate shift problem by a normalization step that fixes the means and variances of layer inputs where the estimations of mean and variance are computed after each mini-batch rather than the entire training set. Suppose the layer to normalize has a $d$ dimensional input, *i.e.*, $\mathbf{x} = [x_1, x_2, ..., x_d]^T$. We first normalize the $k$-th dimension as follows:

$$\hat{x}_k = (x_k - \mu_\mathcal{B})/\sqrt{\delta_\mathcal{B}^2 + \epsilon} \tag{34}$$

where $\mu_\mathcal{B}$ and $\delta_\mathcal{B}^2$ are the mean and variance of mini-batch respectively, and $\epsilon$ is a constant value. To enhance the representation ability, the normalized input $\hat{x}_k$ is further transformed into:

$$y_k = \text{BN}_{\gamma,\beta}(x_k) = \gamma \hat{x}_k + \beta \tag{35}$$

where $\gamma$ and $\beta$ are learned parameters. Batch normalization has many advantages compared with global data normalization. Firstly, it reduces internal covariant shift. Secondly, BN reduces the dependence of gradients on the scale of the parameters or of their initial values, which gives a beneficial effect on the gradient flow through the network. This enables the use of higher learning rate without the risk of divergence. Furthermore, BN regularizes the model, and thus reduces the need for Dropout. Finally, BN makes it possible to use saturating nonlinear activation functions without getting stuck in the saturated model.

*3.6.5. Shortcut Connections*

As mentioned above, the vanishing gradient problem of deep CNNs can be alleviated by normalized initialization [8] and BN [120]. Although these methods successfully prevent deep neural networks from overfitting, they also introduce difficulties in optimizing the networks, resulting in worse performances than shallower networks [56, 104, 105]. Such an optimization problem suffered by deeper CNNs is regarded as the degradation problem.

Inspired by Long Short Term Memory (LSTM) networks [121] which use gate functions to determine how much of a neuron's activation value to transform or just pass through. Srivastava *et al.* [122] propose highway networks which enable the optimization of networks with virtually arbitrary depth. The output of their network is given by:

$$\mathbf{x}_{l+1} = \phi_{l+1}(\mathbf{x}_l, \mathbf{W}_H) \cdot \tau_{l+1}(\mathbf{x}_l, \mathbf{W}_T) + \mathbf{x}_l \cdot (\mathbf{1} - \tau_{l+1}(\mathbf{x}_l, \mathbf{W}_T)) \tag{36}$$

where $\mathbf{x}_l$ and $\mathbf{x}_{l+1}$ correspond to the input and output of $l^{\text{th}}$ highway block, $\tau(\cdot)$ is the transform gate and $\phi(\cdot)$ is usually an affine transformation followed by a non-linear activation function (in general it may take other forms). This gating mechanism forces the layer's inputs and outputs to be of the same size and allows highway networks with tens or hundreds of layers to be trained efficiently. The outputs of gates vary significantly with the input examples, demonstrating that the network does not just learn a fixed structure, but dynamically routes data based on specific examples.

Independently, Residual Nets (ResNets) [12] share the same core idea that works in LSTM units. Instead of employing learnable weights for neuron-specific gating, the shortcut connections in ResNets are not gated and untransformed input is directly propagated to the output which brings fewer parameters. The output of ResNets can be represented as follows:

$$\mathbf{x}_{l+1} = \mathbf{x}_l + f_{l+1}(\mathbf{x}_l, \mathbf{W}_F) \tag{37}$$

where $f_l$ is the weight layer, it can be a composite function of operations such as Convolution, BN, ReLU, or Pooling. With residual block, activation of any deeper unit can be written as the sum of the activation of a shallower unit and a residual function. This also implies that gradients can be directly propagated to shallower units, which makes deep ResNets much easier to be optimized than the original mapping function and more efficient to train very deep nets. This is in contrast to usual feedforward networks, where gradients are essentially a series of matrix-vector products, that may vanish as networks grow deeper.

After the original ResNets, He *et al.* [123] follow up with another preactivation variant of ResNets, where they conduct a set of experiments to show that identity shortcut connections are the easiest for networks to learn. They also find that bringing BN forward performs considerably better than using BN after addition. In their comparisons, the residual net with BN + ReLU pre-activation gets higher accuracies than their previous ResNets [12]. Inspired by [123], Shen *et al.* [124] introduce a weighting factor for the output from the convolutional layer, which gradually introduces the trainable layers. The latest Inception-v4 paper [42] also reports that training is accelerated and performance is improved by using identity skip connections across Inception modules. The original ResNets and preactivation ResNets are very deep but also very thin. By contrast, Wide ResNets [125] proposes to decrease the depth and increase the width, which achieves impressive results on CIFAR-10, CIFAR-100, and SVHN. However, their claims are not validated on the large-scale image classification task on Imagenet dataset[2]. Stochastic Depth ResNets randomly drop a subset of layers and bypass them with the identity mapping for every mini-batch. By combining Stochastic Depth ResNets and Dropout, Singh *et al.* [126] generalize dropout and networks with stochastic depth, which can be viewed as an ensemble of ResNets, Dropout ResNets, and Stochastic Depth ResNets. The ResNets in ResNets (RiR) paper [127] describes an architecture that merges classical convolutional networks and residual networks, where each block of RiR contains residual units and non-residual blocks. The RiR can learn how many convolutional layers it should use per residual block. ResNets of ResNets (RoR) [128] is a modification to the ResNets architecture which proposes to use multi-level shortcut connections as opposed to single-level shortcut connections in the prior work on ResNets [12]. DenseNet [129] can be seen as an architecture takes the insights of the skip connection to the extreme, in which the output of a layer is connected to all the subsequent layer in the module. In all of the ResNets [12, 123], Highway [122] and Inception networks [42], we can see a pretty clear trend of using shortcut connections to help train very deep networks.

## 4. Fast Processing of CNNs

With the increasing challenges in the computer vision and machine learning tasks, the models of deep neural networks get more and more complex. These powerful models require more data for training in order to avoid overfitting. Meanwhile, the big training data also brings new challenges such as how to train the networks in a feasible amount of time. In this section, we introduce some fast processing methods of CNNs.

### 4.1. FFT

Mathieu *et al.* [49] carry out the convolutional operation in the Fourier domain with FFTs. Using FFT-based methods has many advantages. Firstly, the Fourier transformations of filters can be reused as the filters are convolved with multiple images in a mini-batch. Secondly, the Fourier transformations of the output gradients can be reused when backpropagating gradients to both filters and input images. Finally, the summation over input channels can be performed in the Fourier domain, so that inverse Fourier transformations are only required once per output channel per image. There have already been some GPU-based libraries developed to speed up the training and testing process, such as cuDNN [130] and fbfft [131].

---
[2]http://www.image-net.org

However, using FFT to perform convolution needs additional memory to store the feature maps in the Fourier domain, since the filters must be padded to be the same size as the inputs. This is especially costly when the striding parameter is larger than 1, which is common in many state-of-art networks, such as the early layers in [132] and [10]. While FFT can achieve faster training and testing process, the rising prominence of small size convolutional filters have become an important component in CNNs such as ResNet [12] and GoogleNet [10], which makes a new approach specialized for small filter sizes: Winograd's minimal filtering algorithms [133]. The insight of Winograd is like FFT, and the Winograd convolutions can be reduced across channels in transform space before applying the inverse transform and thus makes the inference more efficient.

### 4.2. Structured Transforms

Low-rank matrix factorization has been exploited in a variety of contexts to improve the optimization problems. Given an $m \times n$ matrix $\mathbf{C}$ of rank $r$, there exists a factorization $\mathbf{C} = \mathbf{AB}$ where $\mathbf{A}$ is an $m \times r$ full column rank matrix and $\mathbf{B}$ is an $r \times n$ full row rank matrix. Thus, we can replace $\mathbf{C}$ by $\mathbf{A}$ and $\mathbf{B}$. To reduce the parameters of $\mathbf{C}$ by a fraction $p$, it is essential to ensure that $mr + rn < pmn$, i.e., the rank of $\mathbf{C}$ should satisfy that $r < pmn/(m+n)$. By applying this factorization, the space complexity reduces from $\mathcal{O}(mn)$ to $\mathcal{O}(r(m+n))$, and the time complexity reduces from $\mathcal{O}(mn)$ to $\mathcal{O}(r(m+n))$. To this end, Sainath et al. [134] apply the low-rank matrix factorization to the final weight layer in a deep CNN, resulting about 30-50% speedup in training time with little loss in accuracy. Similarly, Xue et al. [135] apply singular value decomposition on each layer of a deep CNN to reduce the model size by 71% with less than 1% relative accuracy loss. Inspired by [136] which demonstrates the redundancy in the parameters of deep neural networks, Denton et al. [137] and Jaderberg et al. [138] independently investigate the redundancy within the convolutional filters and develop approximations to reduced the required computations. Novikov et al. [139] generalize the low-rank ideas, where they treat the weight matrix as multi-dimensional tensor and apply a Tensor-Train decomposition [140] to reduce the number of parameters of the fully-connected layers.

Adaptive Fastfood transform is generalization of the Fastfood [141] transform for approximating matrix. It reparameterizes the weight matrix $\mathbf{C} \in \mathbb{R}^{n \times n}$ in fully-connected layers with an Adaptive Fastfood transformation: $\mathbf{Cx} = (\mathbf{\tilde{D}_1 H \tilde{D}_2 \Pi H \tilde{D}_3})\mathbf{x}$, where $\mathbf{\tilde{D}}_1$, $\mathbf{\tilde{D}}_2$ and $\mathbf{\tilde{D}}_3$ are diagonal matrices of parameters, $\Pi$ is a random permutation matrix, and $\mathbf{H}$ denotes the Walsh-Hadamard matrix. The space complexity of Adaptive Fastfood transform is $\mathcal{O}(n)$, and the time complexity is $\mathcal{O}(n \log n)$.

Motivated by the great advantages of circulant matrix in both space and computation efficiency [142, 143], Cheng et al. [144] explore the redundancy in the parametrization of fully-connected layers by imposing the circulant structure on the weight matrix to speed up the computation, and further allow the use of FFT for faster computation. With a circulant matrix $\mathbf{C} \in \mathbb{R}^{n \times n}$ as the matrix of parameters in a fully-connected layer, for an input vector $\mathbf{x} \in \mathbb{R}^n$, the output of $\mathbf{Cx}$ can be calculated efficiently using the FFT and inverse IFFT: $\mathbf{CDx} = \text{ifft}(\text{fft}(\mathbf{v})) \circ \text{fft}(\mathbf{x})$, where $\circ$ corresponds to elementwise multiplication operation, $\mathbf{v} \in \mathbb{R}^n$ is defined by $\mathbf{C}$, and $\mathbf{D}$ is a random sign flipping matrix for improving the capacity of the model. This method reduces the time complexity from $\mathcal{O}(n^2)$ to $\mathcal{O}(n \log n)$, and space complexity from $\mathcal{O}(n^2)$ to $\mathcal{O}(n)$. Moczulski et al. [145] further generalize the circulant structures by interleaving diagonal matrices with the orthogonal Discrete Cosine Transform (DCT). The resulting transform, $\mathbf{ACDC}^{-1}$, has $\mathcal{O}(n)$ space complexity and $\mathcal{O}(n \log n)$ time complexity.

### 4.3. Low Precision

Floating point numbers are a natural choice for handling the small updates of the parameters of CNNs. However, the resulting parameters may contain a lot of redundant information [146]. To reduce redundancy, Binarized Neural Networks (BNNs) restrict some or all the arithmetics involved in computing the outputs to be binary values.

There are three aspects of binarization for neural network layers: binary input activations, binary synapse weights, and binary output activations. Full binarization requires all the three components are binarized, and the cases with one or two components are considered as partial binarization. Kim et al. [147] consider full binarization with a predetermined portion of the synapses having zero weight, and all other synapses with a weight of one. Their network only needs XNOR and bit count operations, and they report 98.7% accuracy on the MNIST dataset. XNOR-Net [148] applies convolutional BNNs on the ImageNet dataset with topologies inspired by AlexNet, ResNet and GoogLeNet, reporting top-1 accuracies of up to 51.2% for full binarization and 65.5% for partial binarization. DoReFa-Net [149] explores reducing precision during the forward pass as well as the backward pass. Both partial and full binarization are explored in their experiments and the corresponding top-1 accuracies on ImageNet are 43% and 53%. The work by Courbariaux *et al.* [150] describes how to train fully-connected networks and CNNs with full binarization and batch normalization layers, reporting competitive accuracies on the MNIST, SVHN, and CIFAR-10 datasets.

*4.4. Weight Compression*

Many attempts have been made to reduce the number of parameters in the convolution layers and fully-connected layers. Here, we briefly introduce some methods under these topics: vector quantization, pruning, and hashing.

Vector Quantization (VQ) is a method for compressing densely connected layers to make CNN models smaller. Similar to scalar quantization where a large set of numbers is mapped to a smaller set [151], VQ quantizes groups of numbers together rather than addressing them one at a time. In 2013, Denil *et al.* [136] demonstrate the presence of redundancy in neural network parameters, and use VQ to significantly reduce the number of dynamic parameters in deep models. Gong *et al.* [152] investigate the information theoretical vector quantization methods for compressing the parameters of CNNs, and they obtain parameter prediction results similar to those of [136]. They also find that VQ methods have a clear gain over existing matrix factorization methods, and among the VQ methods, structured quantization methods such as product quantization work significantly better than other methods (*e.g.*, residual quantization [153], scalar quantization [154]).

An alternative approach to weight compression is pruning. It reduces the number of parameters and operations in CNNs by permanently dropping less important connections [155], which enables smaller networks to inherit knowledge from the large predecessor networks and maintains comparable of performance. Han *et al.* [146, 156] introduce fine-grained sparsity in a network by a magnitude-based pruning approach. If the absolute magnitude of any weight is less than a scalar threshold, the weight is pruned. Gao *et al.* [157] extend the magnitude-based approach to allow restoration of the pruned weights in the previous iterations, with tightly coupled pruning and retraining stages, for greater model compression. Yang *et al.* [158] take the correlation between weights into consideration and propose an energy-aware pruning algorithm that directly uses energy consumption estimation of a CNN to guide the pruning process. Rather than fine-grained pruning, there are also works that investigate coarse-grained pruning. Hu *et al.* [159] propose removing filters that frequently generate zero output activations on the validation set. Srinivas *et al.* [160] merge similar filters into one, while Mariet *et al.* [161] merge filters with similar output activations into one.

Designing a proper hashing technique to accelerate the training of CNNs or save memory space also an interesting problem. HashedNets [162] is a recent technique to reduce model sizes by using a hash function to group connection weights into hash buckets, and all connections within the same hash bucket share a single parameter value. Their network shrinks the storage costs of neural networks significantly while mostly preserves the generalization performance in image classification. As pointed out in Shi *et al.* [163] and Weinberger *et al.* [164], sparsity will minimize hash collision making feature hashing even more effective. HashNets may be used together with pruning to give even better parameter savings.

*4.5. Sparse Convolution*

Recently, several attempts have been made to sparsify the weights of convolutional layers [165, 166]. Liu *et al.* [165] consider sparse representations of the basis filters, and achieve 90% sparsifying by exploiting both inter-channel and intra-channel redundancy of convolutional kernels. Instead of sparsifying the weights of convolution layers, Wen *et al.* [166] propose a Structured Sparsity Learning (SSL) approach to simultaneously optimize their hyperparameters (filter size, depth, and local connectivity). Bagherinezhad *et al.* [167] propose a lookup-based convolutional neural network (LCNN) that encodes convolutions by few lookups to a rich set of dictionary that is trained to cover the space of weights in CNNs. They decode the weights of the convolutional layer with a dictionary and two tensors. The dictionary is shared among all weight filters in a layer, which allows a CNN to learn from very few training examples. LCNN can achieve a higher accuracy in a small number of iterations compared to standard CNN.

## 5. Applications of CNNs

In this section, we introduce some recent works that apply CNNs to achieve state-of-the-art performance, including image classification, object tracking, pose estimation, text detection, visual saliency detection, action recognition, scene labeling, speech and natural language processing.

### 5.1. Image Classification

CNNs have been applied in image classification for a long time [168–171]. Compared with other methods, CNNs can achieve better classification accuracy on large scale datasets [8, 9, 172] due to their capability of joint feature and classifier learning. The breakthrough of large scale image classification comes in 2012. Krizhevsky *et al.* [8] develop the AlexNet and achieve the best performance in ILSVRC 2012. After the success of AlexNet, several works have made significant improvements in classification accuracy by either reducing filter size [11] or expanding the network depth [9, 10].

Building a hierarchy of classifiers is a common strategy for image classification with a large number of classes [173]. The work of [174] is one of the earliest attempts to introduce category hierarchy in CNN, in which a discriminative transfer learning with tree-based priors is proposed. They use a hierarchy of classes for sharing information among related classes in order to improve performance for classes with very few training examples. Similarly, Wang *et al.* [175] build a tree structure to learn fine-grained features for subcategory recognition. Xiao *et al.* [176] propose a training method that grows a network not only incrementally but also hierarchically. In their method, classes are grouped according to similarities and are self-organized into different levels. Yan *et al.* [177] introduce a hierarchical deep CNNs (HD-CNNs) by embedding deep CNNs into a category hierarchy. They decompose the classification task into two steps. The coarse category CNN classifier is first used to separate easy classes from each other, and then those more challenging classes are routed downstream to fine category classifiers for further prediction. This architecture follows the coarse-to-fine classification paradigm and can achieve lower error at the cost of an affordable increase of complexity.

Subcategory classification is another rapidly growing subfield of image classification. There are already some fine-grained image datasets (such as Birds [178], Dogs [179], Cars [180], and Plants [181]). Using object part information is beneficial for fine-grained classification [182]. Generally, the accuracy can be improved by localizing important parts of objects and representing their appearances discriminatively. Along this way, Branson *et al.* [183] propose a method which detects parts and extracts CNN features from multiple pose-normalized regions. Part annotation information is used to learn a compact pose normalization space. They also build a model that integrates lower-level feature layers with pose-normalized extraction routines and higher-level feature layers with unaligned image features to improve the classification accuracy. Zhang *et al.* [184] propose a part-based R-CNN which can learn whole-object and part detectors. They use selective search [185] to generate the part proposals, and apply non-parametric geometric constraints to more accurately localize parts. Lin *et al.* [186] incorporate part localization, alignment, and classification into one recognition system which is called Deep LAC. Their system is composed of three sub-networks: localization sub-network is used to estimate the part location, alignment sub-network receives the location as input and performs template alignment [187], and classification sub-network takes pose aligned part images as input to predict the category label. They also propose a value linkage function to link the sub-networks and make them work as a whole in training and testing.

As can be noted, all the above-mentioned methods make use of part annotation information for supervised training. However, these annotations are not easy to collect and these systems have difficulty in scaling up and to handle many types of fine-grained classes. To avoid this problem, some researchers propose to find localized parts or regions in an unsupervised manner. Krause *et al.* [188] use the ensemble of localized learned feature representations for fine-grained classification, they use co-segmentation and alignment to generate parts, and then compare the appearance of each part and aggregate the similarities together. In their latest paper [189], they combine co-segmentation and alignment in a discriminative mixture to generate parts for facilitating fine-grained classification. Zhang *et al.* [190] use the unsupervised selective search to generate object proposals, and then select the useful parts from the multi-scale generated part proposals. Xiao *et al.* [191] apply visual attention in CNN for fine-grained classification. Their classification pipeline is composed of three types of attentions: the bottom-up attention proposes candidate patches, the object-level top-down attention selects relevant patches of a certain object, and the part-level top-down attention localizes discriminative parts. These attentions are combined to train domain-specific networks which can help to find foreground object or object parts and extract discriminative features. Lin *et al.* [192] propose a bilinear model for fine-grained image classification. The recognition architecture consists of two feature extractors. The outputs of two feature extractors are multiplied using the outer product at each location of the image, and are pooled to obtain an image descriptor.

*5.2. Object Detection*

Object detection has been a long-standing and important problem in computer vision [193–195]. Generally, the difficulties mainly lie in how to accurately and efficiently localize objects in images or video frames. The use of CNNs for detection and localization can be traced back to 1990s [196]. However, due to the lack of training data and limited processing resources, the progress of CNN-based object detection is slow before 2012. Since 2012, the huge success of CNNs in ImageNet challenge [8] rekindles interest in CNN-based object detection [197]. In some early works [196, 198], they use the sliding window based approaches to densely evaluate the CNN classifier on windows sampled at each location and scale. Since there are usually hundreds of thousands of candidate windows in a image, these methods suffer from highly computational cost, which makes them unsuitable to be applied on the large-scale dataset, *e.g.*, Pascal VOC [172], ImageNet [8] and MSCOCO [199].

Recently, object proposal based methods attract a lot of interests and are widely studied in the literature [185, 193, 200, 201]. These methods usually exploit fast and generic measurements to test whether a sampled window is a potential object or not, and further pass the output object proposals to more sophisticated detectors to determine whether they are background or belong to a specific object class. One of the most famous object proposal based CNN detector is Region-based CNN (R-CNN) [202]. R-CNN uses Selective Search (SS) [185] to extract around 2000 bottom-up region proposals that are likely to contain objects. Then, these region proposals are warped to a fixed size ($227 \times 227$), and a pre-trained CNN is used to extract features from them. Finally, a binary SVM classifier is used for detection.

R-CNN yields a significant performance boost. However, its computational cost is still high since the time-consuming CNN feature extractor will be performed for each region separately. To deal with this problem, some recent works propose to share the computation in feature extraction [9, 28, 132, 202]. OverFeat [132] computes CNN features from an image pyramid for localization and detection. Hence the computation can be easily shared between overlapping windows. Spatial pyramid pooling network (SPP net) [203] is a pyramid-based version of R-CNN [202], which introduces an SPP layer to relax the constraint that input images must have a fixed size. Unlike R-CNN, SPP net extracts the feature maps from the entire image only once, and then applies spatial pyramid pooling on each candidate window to get a fixed-length representation. A drawback of SPP net is that its training procedure is a multi-stage pipeline, which makes it impossible to train the CNN feature extractor and SVM classifier jointly to further improve the accuracy. Fast RCNN [204] improves SPP net by using an end-to-end training method. All network layers can be updated during fine-tuning, which simplifies the learning process and improves detection accuracy. Later, Faster R-CNN [204] introduces a region proposal network (RPN) for object proposals generation and achieves further speed-up. Beside R-CNN based methods, Gidaris *et al.* [205] propose a multi-region and semantic segmentation-aware model for object detection. They integrate the combined features on an iterative localization mechanism as well as a box-voting scheme after non-max suppression. Yoo *et al.* [206] treat the object detection problem as an iterative classification problem. It predicts an accurate object boundary box by aggregating quantized weak directions from their detection network.

Another important issue of object detection is how to explore effective training sets as the performance is somehow largely depends on quantity and quality of both positive and negative samples. Online bootstrap-ping (or hard negative mining [207]) for CNN training has recently gained interest due to its importance for intelligent cognitive systems interacting with dynamically changing environments [208]. [209] proposes a novel bootstrapping technique called online hard example mining (OHEM) for training detection models based on CNNs. It simplifies the training process by automatically selecting the hard examples. Meanwhile, it only computes the feature maps of an image once, and then forwards all region-of-interests (RoIs) of the image on top of these feature maps. Thus it is able to find the hard examples with a small extra computational cost.

More recently, YOLO [210] and SSD [211] allow single pipeline detection that directly predicts class labels. YOLO [210] treats object detection as a regression problem to spatially separated bounding boxes and associated class probabilities. The whole detection pipeline is a single network which predicts bounding boxes and class probabilities from the full image in one evaluation, and can be optimized end-to-end directly on detection performance. SSD [211] discretizes the output space of bounding boxes into a set of default boxes over different aspect ratios and scales per feature map location. With this multiple scales setting and their matching strategy, SSD is significantly more accurate than YOLO. With the benefits from super-resolution, Lu *et al.* [212] propose a top-down search strategy to divide a window into sub-windows recursively, in which an additional network is trained to account for such division decisions.

### 5.3. Object Tracking

The success in object tracking relies heavily on how robust the representation of target appearance is against several challenges such as view point changes, illumination changes, and occlusions [213–215]. There are several attempts to employ CNNs for visual tracking. Fan *et al.* [216] use CNN as a base learner. It learns a separate class-specific network to track objects. In [216], the authors design a CNN tracker with a shift-variant architecture. Such an architecture plays a key role so that it turns the CNN model from a detector into a tracker. The features are learned during offline training. Different from traditional trackers which only extract local spatial structures, this CNN based tracking method extracts both spatial and temporal structures by considering the images of two consecutive frames. Because the large signals in the temporal information tend to occur near objects that are moving, the temporal structures provide a crude velocity signal to tracking.

Li *et al.* [217] propose a target-specific CNN for object tracking, where the CNN is trained incrementally during tracking with new examples obtained online. They employ a candidate pool of multiple CNNs as a data-driven model of different instances of the target object. Individually, each CNN maintains a specific set of kernels that favourably discriminate object patches from their surrounding background using all available low-level cues. These kernels are updated in an online manner at each frame after being trained with just one instance at the initialization of the corresponding CNN. Instead of learning one complicated and powerful CNN model for all the appearance observations in the past, Li *et al.* [217] use a relatively small number of filters in the CNN within a framework equipped with a temporal adaptation mechanism. Given a frame, the most promising CNNs in the pool are selected to evaluate the hypothesises for the target object. The hypothesis with the highest score is assigned as the current detection window and the selected models are retrained using a warm-start backpropagation which optimizes a structural loss function.

In [218], a CNN object tracking method is proposed to address limitations of handcrafted features and shallow classifier structures in object tracking problem. The discriminative features are first automatically learned via a CNN. To alleviate the tracker drifting problem caused by model update, the tracker exploits the ground truth appearance information of the object labeled in the initial frames and the image observations obtained online. A heuristic schema is used to judge whether updating the object appearance models or not.

Hong *et al.* [219] propose a visual tracking algorithm based on a pre-trained CNN, where the network is trained originally for large-scale image classification and the learned representation is transferred to describe target. On top of the hidden layers in the CNN, they put an additional layer of an online SVM to learn a target appearance discriminatively against background. The model learned by SVM is used to compute a target-specific saliency map by back-projecting the information relevant to target to input image space. And they exploit the target-specific saliency map to obtain generative target appearance models and perform tracking with understanding of spatial configuration of target.

*5.4. Pose Estimation*

Since the breakthrough in deep structure learning, many recent works pay more attention to learn multiple levels of representations and abstractions for human-body pose estimation task with CNNs [220, 221]. DeepPose [222] is the first application of CNNs to human pose estimation problem. In this work, pose estimation is formulated as a CNN-based regression problem to body joint coordinates. A cascade of 7-layered CNNs are presented to reason about pose in a holistic manner. Unlike the previous works that usually explicitly design graphical model and part detectors, the DeepPose captures the full context of each body joint by taking the whole image as the input.

Meanwhile, some works exploit CNN to learn representation of local body parts. Ajrun *et al.* [223] present a CNN based end-to-end learning approach for full-body human pose estimation, in which CNN part detectors and an Markov Random Field (MRF)-like spatial model are jointly trained, and pair-wise potentials in the graph are computed using convolutional priors. In a series of papers, Tompson *et al.* [224] use a multi-resolution CNN to compute heat-map for each body part. Different from [223], Tompson *et al.* [224] learn the body part prior model and implicitly the structure of the spatial model. Specifically, they start by connecting every body part to itself and to every other body part in a pair-wise fashion, and use a fully-connected graph to model the spatial prior. As an extension of [224], Tompson *et al.* [92] propose a CNN architecture which includes a position refinement model after a rough pose estimation CNN. This refinement model, which is a Siamese network [64], is jointly trained in cascade with the off-the-shelf model [224]. In a similar work with [224], Chen *et al.* [225, 226] also combine graphical model with CNN. They exploit a CNN to learn conditional probabilities for the presence of parts and their spatial relationships, which are used in unary and pairwise terms of the graphical model. The learned conditional probabilities can be regarded as low-dimensional representations of the body pose. There is also a pose estimation method called dual-source CNN [227] that integrates graphical models and holistic style. It takes the full body image and the holistic view of the local parts as inputs to combine both local and contextual information.

In addition to still image pose estimation with CNN, recently researchers also apply CNN to human pose estimation in videos. Based on the work [224], Jain *et al.* [228] also incorporate RGB features and motion features to a multi-resolution CNN architecture to further improve accuracy. Specifically, The CNN works in a sliding-window manner to perform pose estimation. The input of the CNN is a 3D tensor which consists of an RGB image and its corresponding motion features, and the output is a 3D tensor containing response-maps of the joints. In each response map, the value of each location denote the energy for presence the corresponding joint at that pixel location. The multi-resolution processing is achieved by simply down sampling the inputs and feeding them to the network.

*5.5. Text Detection and Recognition*

The task of recognizing text in image has been widely studied for a long time [229–232]. Traditionally, optical character recognition (OCR) is the major focus. OCR techniques mainly perform text recognition on images in rather constrained visual environments (*e.g.*, clean background, well-aligned text). Recently, the focus has been shifted to text recognition on scene images due to the growing trend of high-level visual understanding in computer vision research [233, 234]. The scene images are captured in unconstrained environments where there exists a large amount of appearance variations which poses great difficulties to existing OCR techniques. Such a concern can be mitigated by using stronger and richer feature representations such as those learned by CNN models. Along the line of improving the performance of scene text recognition with CNN, a few works have been proposed. The works can be coarsely categorized into three types: (1) text detection and localization without recognition, (2) text recognition on cropped text images, and (3) end-to-end text spotting that integrates both text detection and recognition:

*5.5.1. Text Detection*

One of the pioneering works to apply CNN for scene text detection is [235]. The CNN model employed by [235] learns on cropped text patches and non-text scene patches to discriminate between the two. The text are then detected on the response maps generated by the CNN filters given the multiscale image pyramid of the input. To reduce the search space for text detection, Xu *et al.* [236] propose to obtain a set of character candidates via Maximally Stable Extremal Regions (MSER) and filter the candidates by CNN classification. Another work that combines MSER and CNN for text detection is [237]. In [237], CNN is used to distinguish text-like MSER components from non-text components, and cluttered text components are split by applying CNN in a sliding window manner followed by Non-Maximal Suppression (NMS). Other than localization of text, there is an interesting work [238] that makes use of CNN to determine whether the input image contains text, without telling where the text is exactly located. In [238], text candidates are obtained using MSER which are then passed into a CNN to generate visual features, and lastly the global features of the images are constructed by aggregating the CNN features in a Bag-of-Words (BoW) framework.

*5.5.2. Text Recognition*

Goodfellow *et al.* [239] propose a CNN model with multiple softmax classifiers in its final layer, which is formulated in such a way that each classifier is responsible for character prediction at each sequential location in the multi-digit input image. As an attempt to recognize text without using lexicon and dictionary, Jaderberg *et al.* [240] introduce a novel Conditional Random Fields (CRF)-like CNN model to jointly learn character sequence prediction and bigram generation for scene text recognition. The more recent text recognition methods supplement conventional CNN models with variants of recurrent neural networks (RNN) to better model the sequence dependencies between characters in text. In [241], CNN extracts rich visual features from character-level image patches obtained via sliding window, and the sequence labelling is carried out by LSTM [242]. The method presented in [243] is very similar to [241], except that in [243], lexicon can be taken into consideration to enhance text recognition performance.

*5.5.3. End-to-end Text Spotting*

For end-to-end text spotting, Wang *et al.* [15] apply a CNN model originally trained for character classification to perform text detection. Going in a similar direction as [15], the CNN model proposed in [244] enables feature sharing across the four different subtasks of an end-to-end text spotting system: text detection, character case-sensitive and insensitive classification, and bigram classification. Jaderberg *et al.* [245] make use of CNNs in a very comprehensive way to perform end-to-end text spotting. In [245], the major subtasks of its proposed system, namely text bounding box filtering, text bounding box regression, and text recognition are each tackled by a separate CNN model.

*5.6. Visual Saliency Detection*

The technique to locate important regions in imagery is referred to as visual saliency prediction. It is a challenging research topic, with a vast number of computer vision and image processing applications facilitated by it. Recently, a couple of works have been proposed to harness the strong visual modeling capability of CNNs for visual saliency prediction.

Multi-contextual information is a crucial prior in visual saliency prediction, and it has been used concurrently with CNN in most of the considered works [246–250]. Wang *et al.* [246] introduce a novel saliency detection algorithm which sequentially exploits local context and global context. The local context is handled by a CNN model which assigns a local saliency value to each pixel given the input of local image patches, while the global context (object-level information) is handled by a deep fully-connected feedforward network. In [247], the CNN parameters are shared between the global-context and local-context models, for predicting the saliency of superpixels found within object proposals. The CNN model adopted in [248] is pre-trained on large-scale image classification dataset and then shared among different contextual levels for feature extraction. The outputs of the CNN at different contextual levels are then concatenated as input to be passed into a trainable fully-connected feedforward network for saliency prediction. Similar to [247, 248], the CNN model used in [249] for saliency prediction are shared across three CNN streams, with each stream taking input of a different contextual scale. He *et al.* [250] derive a spatial kernel and a range kernel to produce two meaningful sequences as 1-D CNN inputs, to describe color uniqueness and color distribution respectively. The proposed sequences are advantageous over inputs of raw image pixels because they can reduce the training complexity of CNN, while being able to encode the contextual information among superpixels.

There are also CNN-based saliency prediction approaches [251–253] that do not consider multi-contextual information. Instead, they rely very much on the powerful representation capability of CNN. In [251], an ensemble of CNNs is derived from a large number of randomly instantiated CNN models, to generate good features for saliency detection. The CNN models instantiated in [251] are however not deep enough because the maximum number of layers is capped at three. By using a pre-trained and deeper CNN model with 5 convolutional layers, [252] (Deep Gaze) learns a separate saliency model to jointly combine the responses from every CNN layer and predict saliency values. [253] is the only work making use of CNN to perform visual saliency prediction in an end-to-end manner, which means the CNN model accepts raw pixels as input and produces saliency map as output. Pan *et al.* [253] argue that the success of the proposed end-to-end method is attributed to its not-so-deep CNN architecture which attempts to prevent overfitting.

### 5.7. Action Recognition

Action recognition, the behaviour analysis of human subjects and classifying their activities based on their visual appearance and motion dynamics, is one of the challenging problems in computer vision [254–256]. Generally, this problem can be divided to two major groups: action analysis in still images and in videos. For both of these two groups, effective CNN based methods have been proposed. In this subsection we briefly introduce the latest advances on these two groups.

#### 5.7.1. Action Recognition in Still Images

The work of [257] has shown the output of last few layers of a trained CNN can be used as a general visual feature descriptor for a variety of tasks. The same intuition is utilized for action recognition by [9, 258], in which they use the outputs of the penultimate layer of a pre-trained CNN to represent full images of actions as well as the human bounding boxes inside them, and achieve a high level of performance in action classification. Gkioxari *et al.* [259] add a part detection to this framework. Their part detector is a CNN based extension to the original Poselet [260] method.

CNN based representation of contextual information is utilized for action recognition in [261]. They search for the most representative secondary region within a large number of object proposal regions in the image and add contextual features to the description of the primary region (ground truth bounding box of human subject) in a bottom-up manner. They utilize a CNN to represent and fine-tune the representations of the primary and the contextual regions. After that, they move a step forward and show that it is possible to locate and recognize human actions in images without using human bounding boxes [262]. However, they need to train human detectors to guide their recognition at test time. In [263], they propose a method that segments out the action mask of underlying human-object interactions with minimum annotation efforts.

#### 5.7.2. Action Recognition in Video Sequences

Applying CNNs on videos is challenging because traditional CNNs are designed to represent two dimensional pure spatial signals but in videos a new temporal axis is added which is essentially different from the spatial variations in images [256, 264]. The sizes of the video signals are also in higher orders in comparison to those of images which makes it more difficult to apply convolutional networks on. Ji *et al.* [265] propose to consider the temporal axis in a similar manner as other spatial axes and introduce a network of 3D convolutional layers to be applied on video inputs. Recently Tran *et al.* [266] study the performance, efficiency, and effectiveness of this approach and show its strengths compared to other approaches.

Another approach to apply CNNs on videos is to keep the convolutions in 2D and fuse the feature maps of consecutive frames, as proposed by [267]. They evaluate three different fusion policies: late fusion, early fusion, and slow fusion, and compare them with applying the CNN on individual single frames. One more step forward for better action recognition via CNNs is to separate the representation to spatial and temporal variations and train individual CNNs for each of them, as proposed by Simonyan and Zisserman [268]. First stream of this framework is a traditional CNN applied on all the frames and the second receives the dense optical flow of the input videos and trains another CNN which is identical to the spatial stream in size and structure. The output of the two streams are combined in a class score fusion step. Chéron *et al.* [269] utilize the two stream CNN on the localized parts of the human body and show the aggregation of part-based local CNN descriptors can effectively improve the performance of action recognition. Another approach to model the dynamics of videos differently from spatial variations, is to feed the CNN based features of individual frames to a sequence learning module *e.g.*, a recurrent neural network. Donahue *et al.* [270] study different configurations of applying LSTM units as the sequence learner in this framework.

*5.8. Scene Labeling*

Scene labeling aims to relate one semantic class (road, water, sea *etc.*) to each pixel of the input image [271–275]. CNNs are used to model the class likelihood of pixels directly from local image patches. They are able to learn strong features and classifiers to discriminate the local visual subtleties. Farabet *et al.* [276] have pioneered to apply CNNs to scene labeling tasks. They feed their Multi-scale ConvNet with different scale image patches, and they show that the learned network is able to perform much better than systems with hand-crafted features. Besides, this network is also successfully applied to RGB-D scene labeling [277]. To enable the CNNs to have a large field of view over pixels, Pinheiro *et al.* [278] develop the recurrent CNNs. More specifically, the identical CNNs are applied recurrently to the output maps of CNNs in the previous iterations. By doing this, they can achieve slightly better labeling results while significantly reduces the inference times. Shuai *et al.* [279–281] train the parametric CNNs by sampling image patches, which speeds up the training time dramatically. They find that patch-based CNNs suffer from local ambiguity problems, and [279] solve it by integrating global beliefs. [280] and [281] use the recurrent neural networks to model the contextual dependencies among image features from CNNs, and dramatically boost the labeling performance.

Meanwhile, researchers are exploiting to use the pre-trained deep CNNs for object semantic segmentation. Mostajabi *et al.* [282] apply the local and proximal features from a ConvNet and apply the Alex-net [8] to obtain the distant and global features, and their concatenation gives rise to the zoom-out features. They achieve very competitive results on the semantic segmentation tasks. Long *et al.* [28] train a fully convolutional Network to directly predict the input images to dense label maps. The convolution layers of the FCNs are initialized from the model pre-trained on ImageNet classification dataset, and the deconvolution layers are learned to upsample the resolution of label maps. Chen *et al.* [283] also apply the pre-trained deep CNNs to emit the labels of pixels. Considering that the imperfectness of boundary alignment, they further use fully connected CRF to boost the labeling performance.

*5.9. Speech Processing*

*5.9.1. Automatic Speech Recognition*

Automatic Speech Recognition (ASR) is the technology that converts human speech into spoken words [284]. Before applying CNN to ASR, this domain has long been dominated by the Hidden Markov Model and Gaussian Mixture Model (GMM-HMM) methods [285], which usually require extracting hand-craft features on speech signals, *e.g.*, the most popular Mel Frequency Cepstral Coefficients (MFCC) features. Meanwhile, some researchers have applied Deep Neural Networks (DNNs) in large vocabulary continuous speech recognition (LVCSR) and obtained encouraging results [286, 287], however, their networks are susceptible to performance degradations under mismatched condition [288], such as different recording conditions *etc.*

CNNs have shown better performance over GMM-HMMs and general DNNs [289, 290], since they are well suited to exploit the correlations in both time and frequency domains through the local connectivity and are capable of capturing frequency shift in human speech signals. In [289], they achieve lower speech recognition errors by applying CNN on Mel filter bank features. Some attempts use the raw waveform with CNNs, and to learn filters to process the raw waveform jointly with the rest of the network [291, 292]. Most of the early applications of CNN in ASR only use fewer convolution layers. For example, Abdel-Hamid *et al.* [290] use one convolutional layer in their network, and Amodei *et al.* [293] use three convolutional layers as the feature preprocessing layer. Recently, very deep CNNs have shown impressive performance in ASR [294, 295]. Besides, small filters have successfully applied in acoustic modeling in hybrid NN-HMM ASR system, and pooling operations are replaced by densely connected layers for ASR tasks [296]. Yu *et al.* [297] propose a layer-wise context expansion with attention model for ASR. It is a variant of time-delay neural network [298] in which lower layers focus on extracting simple local patterns while higher layers exploit broader context and extract complex patterns than the lower layers. A similar idea can be found in [40].

*5.9.2. Statistical Parametric Speech Synthesis*

In addition to speech recognition, the impact of CNNs has also spread to Statistical Parametric Speech Synthesis (SPSS). The goal of speech synthesis is to generate speech sounds directly from the text and possibly with additional information. It has been known for many years that the speech sounds generated by shallow structured HMM networks are often muffled compared with natural speech. Many studies have adopted deep learning to overcome such deficiency [299–301]. One advantage of these methods is their strong ability to represent the intrinsic correlation by using a generative modeling framework. Inspired by the recent advances in neural autoregressive generative models that model complex distributions such as images [302] and text [303], WaveNet [39] makes use of the generative model of the CNN to represent the conditional distribution of the acoustic features given the linguistic features, which can be seen as a milestone in SPSS. In order to deal with long-range temporal dependencies, they develop a new architecture based on dilated causal convolutions to capture very large receptive fields. By conditioning linguistic features on text, it can be used to directly synthesize speech from text.

*5.10. Natural Language Processing*

*5.10.1. Statistical Language Modeling*

For statistical language modeling, the input typically consists of incomplete sequences of words rather than complete sentences [304, 305]. Kim *et al.* [304] use the output of character-level CNN as the input to an LSTM at each time step. *gen*CNN [306] is a convolutional architecture for sequence prediction, which uses separate gating networks to replace the max-pooling operations. Recently, Kalchbrenner *et al.* [38] propose a CNN-based architecture for sequence processing called ByteNet, which is a stack of two dilated CNNs. Like WaveNet [39], ByteNet also benefits from convolutions with dilations to increase the receptive field size, thus can model sequential data with long-term dependencies. It also has the advantage that the computational time only linearly depends on the length of the sequences. Compared with recurrent neural networks, CNNs not only can get long-range information but also get a hierarchical representation of the input words. Gu *et al.* [307] and Yann *et al.* [308] share a similar idea that both of them use CNN without pooling to model the input words. Gu *et al.* [307] combine the language CNN with recurrent highway networks and achieve a huge improvement compared to LSTM-based methods. Inspired by the gating mechanism in LSTM networks, the gated CNN in [308] uses a gating mechanism to control the path through which information flows in the network, and achieves the state-of-the-art on WiKiText-103. However, the frameworks in [308] and [307] are still under the recurrent framework, and the input window size of their network are of limited size. How to capture the specific long-term dependencies as well as hierarchical representation of history words is still an open problem.

*5.10.2. Text Classification*

Text classification is a crucial task for Natural Language Processing (NLP). Natural language sentences have complicated structures, both sequential and hierarchical, that are essential for understanding them. Owing to the powerful capability of capturing local relations of temporal or hierarchical structures, CNNs have achieved top performance in sentence modeling. A proper CNN architecture is important for text classification. Collobert *et al.* [309] and Yu *et al.* [310] apply one convolutional layer to model the sentence, while Kalchbrenner *et al.* [311] stack multiple layers of convolution to model sentences. In [312], they use multichannel convolution and variable kernels for sentence classification. It is shown that multiple convolutional layers help to extract high-level abstract features, and multiple linear filters can effectively consider different $n$-gram features. Recently, Yin *et al.* [313] extend the network in [312] by hierarchical convolution architecture and further exploration of multichannel and variable size feature detectors. The pooling operation can help the network deal with variable sentence lengths. In [312, 314], they use max-pooling to keep the most important information to represent the sentence. However, max-pooling cannot distinguish whether a relevant feature in one of the rows occurs just one or multiple times and it ignores the order in which the features occur. In [311], they propose the $k$-max pooling which returns the top $k$ activations in the original order in the input sequence. Dynamic $k$-max pooling is a generalization of the $k$-max pooling operator where the $k$ value is depended on the input feature map size. The CNN architectures mentioned above are rather shallow compared with the deep CNNs which are very successful in computer vision. Recently, Conneau *et al.* [315] implement a deep convolutional architecture which is up to 29 convolutional layers. They find that shortcut connections give better results when the network is very deep (49 layers). However, they do not achieve state-of-the-art under this setting.

## 6. Conclusions and Outlook

Deep CNNs have made breakthroughs in processing image, video, speech and text. In this paper, we have given an extensive survey on recent advances of CNNs. We have discussed the improvements of CNN on different aspects, namely, layer design, activation function, loss function, regularization, optimization and fast computation. Beyond surveying the advances of each aspect of CNN, we have also introduced the application of CNN on many tasks, including image classification, object detection, object tracking, pose estimation, text detection, visual saliency detection, action recognition, scene labeling, speech and natural language processing.

Although CNNs have achieved great success in experimental evaluations, there are still lots of issues that deserve further investigation. Firstly, since the recent CNNs are becoming deeper and deeper, they require large-scale dataset and massive computing power for training. Manually collecting labeled dataset requires huge amounts of human efforts. Thus, it is desired to explore unsupervised learning of CNNs. Meanwhile, to speed up training procedure, although there are already some asynchronous SGD algorithms which have shown promising result by using CPU and GPU clusters, it is still worth to develop effective and scalable parallel training algorithms. At testing time, these deep models are highly memory demanding and time-consuming, which makes them not suitable to be deployed on mobile platforms that have limited resources. It is important to investigate how to reduce the complexity and obtain fast-to-execute models without loss of accuracy.

Furthermore, one major barrier for applying CNN on a new task is that it requires considerable skill and experience to select suitable hyperparameters such as the learning rate, kernel sizes of convolutional filters, the number of layers *etc*. These hyper-parameters have internal dependencies which make them particularly expensive for tuning. Recent works have shown that there exists a big room to improve current optimization techniques for learning deep CNN architectures [12, 42, 316].

Finally, the solid theory of CNNs is still lacking. Current CNN model works very well for various applications. However, we do not even know why and how it works essentially. It is desirable to make more efforts on investigating the fundamental principles of CNNs. Meanwhile, it is also worth exploring how to leverage natural visual perception mechanism to further improve the design of CNN. We hope that this paper not only provides a better understanding of CNNs but also facilitates future research activities and application developments in the field of CNNs.

**Acknowledgment**

This research was carried out at the Rapid-Rich Object Search (ROSE) Lab at the Nanyang Technological University, Singapore. The ROSE Lab is supported by the Infocomm Media Development Authority, Singapore.